\documentclass[journal]{IEEEtran}
\ifCLASSINFOpdf
  \usepackage[pdftex]{graphicx}
\else
\fi
\usepackage{amsmath}
\usepackage{amssymb}

\usepackage{mathtools}

\DeclareMathOperator*{\argmax}{argmax}

\ifCLASSOPTIONcompsoc
 \usepackage[caption=false,font=normalsize,labelfont=sf,textfont=sf]{subfig}
\else
 \usepackage[caption=false,font=footnotesize]{subfig}
\fi

\usepackage{amsmath,amsfonts,amssymb}
\usepackage{amsthm}  {

      \newtheorem*{example*}{Example}

}

\usepackage{booktabs,tabularx}
\newcolumntype{L}{>{\raggedright\arraybackslash}X}

\begin{document}
\title{Partially Observable Markov Decision Processes in
Robotics: A Survey}

\author{{Mikko Lauri, David Hsu and Joni Pajarinen}%
\thanks{M.~Lauri is with the Department of Informatics, Universit\"{a}t Hamburg, Germany (e-mail: {\tt\footnotesize mikko.lauri@uni-hamburg.de}).}%
\thanks{D.~Hsu is with the Department of Computer Science, National University of Singapore, Singapore (e-mail: {\tt\footnotesize dyhsu@comp.nus.edu.sg}).}%
\thanks{J.~Pajarinen is with the Department of Electrical Engineering and Automation, Aalto University, Finland and Intelligent Autonomous Systems, Technische Universit\"{a}t Darmstadt, Germany (e-mail: {\tt\footnotesize joni.pajarinen@aalto.fi}).}%
}

\markboth{IEEE Transactions on Robotics}%
{X \MakeLowercase{\textit{et al.}}: POMDP Survey}
\maketitle

\begin{abstract}
Noisy sensing, imperfect control, and environment changes are defining characteristics of many real-world robot tasks.
The \textit{partially observable Markov decision process} (POMDP) provides a principled mathematical framework for modeling and solving robot decision and control  tasks under uncertainty.
Over the last decade, it has seen many successful applications, spanning localization and navigation, search and tracking, autonomous driving, multi-robot systems, manipulation, and human-robot interaction. 
This survey aims to bridge the gap between the development of POMDP models and algorithms at one end and application to diverse robot decision tasks at the other. 
It analyzes the characteristics of these tasks and connects them with the mathematical and algorithmic properties of the POMDP framework for effective modeling and solution. 
For practitioners, the survey provides some of the key task characteristics in deciding when and how to apply POMDPs to robot tasks successfully.  
For POMDP algorithm designers, the survey provides new insights into the unique challenges of applying POMDPs to robot systems and points to promising new  directions for further research.
\end{abstract}


\IEEEpeerreviewmaketitle

\section{Introduction}

\IEEEPARstart{U}{ncertainties} are ubiquitous in robot systems {due to} noisy sensing, imperfect robot control,  fast-changing environments, and inaccurate models. 
A robot must  reason about the possible outcomes
of its actions based on limited sensor information; it
takes actions that not only yield short-term reward but also  gather information
 for long-term success. For example, an autonomous robot vehicle tries to pass through an unsignalized traffic intersection as fast as possible. Instead of accelerating, the vehicle may  have to slow down in the short term, in order to gather information on the intentions of pedestrians and other vehicles. This  information helps the vehicle to coordinate its actions with others and achieve the overall goal faster  in the long term~\cite{bai2015intention,brechtel2014probabilistic}. 
Similarly, a robot manipulator tries to push an irregular object to a designated pose, with the minimum number of actions. Instead of pushing the objects directly towards the final pose, it may use the first few pushes to gather information on the object's center of mass so that the later pushes become much more effective~\cite{koval2016pre}.
The \emph{partially observable Markov decision process} (POMDP)~\cite{astrom1965optimal} is a principled general  framework for such robot decision-making tasks under uncertainty.

A POMDP models  a decision-maker, also called the \emph{agent}, for a system with {incomplete state information}.
At each  time step, the agent executes an action that yields {some} reward {depending} on the current state and the action, and results in a stochastic transition to a new {system} state.
The agent acquires information about the new state via noisy observations.
A solution to a POMDP is a \emph{policy} that prescribes the agent's action  conditioned on its past history of actions and observations.
The performance of a policy is measured by an objective function, which measures the expected total discounted {reward} over time.
An optimal policy maximizes this objective function.
To compute an optimal policy,  \emph{planning algorithms} reason about the possible long-term effects of actions on future states, observations, and { reward using state transition, observation, and reward models}.
{A prerequisite for planning is to acquire these models.
For some robotic tasks, the models {are} specified by a domain expert, or they {are} learned from data prior to planning.
The model-based POMDP planning approach therefore differs from other control and learning approaches, such as reinforcement learning~(e.g.,~\cite{kober2013reinforcement}), which do not use system models or sometimes learn such models concurrently with the policy by interacting with the system.}

POMDP planning {offers key conceptual, algorithmic, and practical advantages for robot systems.}
First, {The POMDP models uncertainties inherent to robot} systems. A robust robot system must reason  about the effects of noisy sensor observations, imperfect robot actions, and environment changes. This is unavoidable. The POMDP provides a principled conceptual framework to model these uncertainties probabilistically and to identify the underlying conditions required for an optimal solution. The POMDP framework is general and applicable to most  robot systems commonly encountered.
Second, while POMDP planning is computationally intractable in the worst case, there have been dramatic successes in developing  efficient \textit{approximation algorithms} through Monte Carlo sampling and simulation (Section~\ref{sec:solvers}).  Given the POMDP model of a robot system, these algorithms solve for a near-optimal policy automatically, reducing the work required to manually engineer such policies and  improving the practical performance of robot systems under uncertainty (Section~\ref{sec:applications}). 
Finally, some POMDP planning algorithms provide performance guarantees~\cite{smith2005point,pineau2006anytime,somani2013despot}, {\cite{thiebaux2016rao}}, a key requirement for safety-critical applications.

{There are several surveys on planning algorithms for POMDPs in the operations research and artificial intelligence research communities}~\cite{hauskrecht2000value,monahan1982state,white1991survey,lovejoy1991survey,shani2013survey}.
Likewise, there are many surveys covering robotics applications such as grasping~\cite{bohg2014grasping}, human robot interaction~\cite{lasota2017safe}, and autonomous driving~\cite{sorin2019deep}.
What is lacking, however, is a synthesised view of how to apply the POMDP effectively across different {robot systems}.

This survey covers the application of POMDP planning to robotics in six broad categories:
\begin{itemize}
    \item localization and navigation
    \item autonomous driving
    \item target search, tracking and avoidance
    \item manipulation and grasping
    \item human-robot interaction
    \item multi-robot coordination.
\end{itemize}
We identify general challenges of applying POMDPs to {robotic} tasks,  including dealing with high-dimensional, continuous state, action, and observations spaces, and designing accurate and useful models for planning.
For practitioners interested in the POMDP as a tool for robot decision making under uncertainty,  the survey provides insights in a range of applications and their key characteristics that enable effective POMDP solutions. 
For researchers specializing in POMDP planning algorithms, this survey serves as an overview of the unique challenges in applying POMDPs to {robotic systems} and suggests potential directions for future research.

The remainder of this survey is organized as follows.
In Section~\ref{sec:pomdps}, we  summarize the theoretical foundations of the POMDP model.
In Section~\ref{sec:solvers}, we provide a short review of solution algorithms for computing (near-)optimal policies for POMDPs.
Section~\ref{sec:applications} presents the POMDP model and solution algorithms for the six {robotic tasks} listed above. 
In Section~\ref{sec:open_questions}, we discuss future challenges  most important for the application of POMDPs to robot tasks. 
Section~\ref{sec:conclusion} ends with some concluding remarks.

\section{POMDP Models}
\label{sec:pomdps}
In this section, we review the POMDP model for the finite horizon and infinite horizon cases.
{We defer discussion of practical solution algorithms to the next section.}
We assume the reader is familiar with the concept of recursive state estimation or Bayesian filtering{~\cite{chen2003bayesian}}.
For simplicity, we only consider the case where the sets of states, actions, and observations are finite, and the objective function is the expected {total discounted reward}.
We refer the reader interested in further theoretical background to~\cite{puterman1994markov,kaelbling1998planning,hauskrecht2000value}.

\subsection{Finite horizon}
\label{subsec:finitehorizon}
\begin{figure}[!t]
    \centering
    \includegraphics[width=0.7\columnwidth]{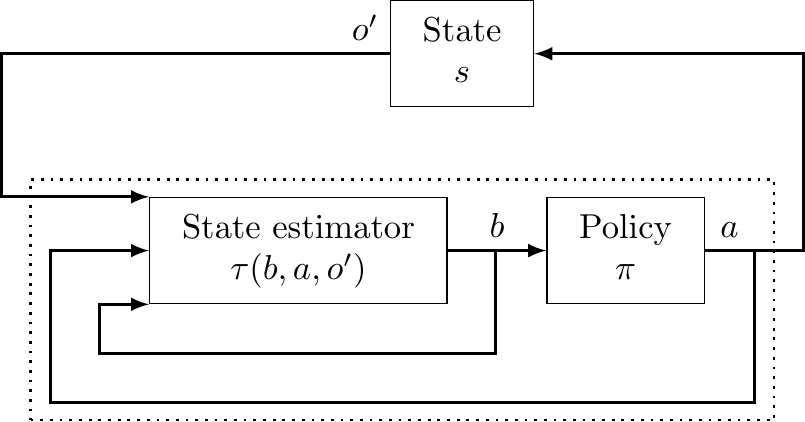}
    \caption{A single interaction step of an agent (dashed box) in a robot system with a partially observed state $s$. {Note that the state may include both the external environmental state and the internal state of the robot system.} An action $a$ is selected according to a policy $\pi$ as a function of the current {belief} $b$. The action causes a transition from $s$ to $s'$, perceived by the agent via an observation $o'$. Finally, $b$, $a$, and $o'$ are used to update the {belief} using a Bayes filter $\tau$.}
    \label{fig:pomdp}
\end{figure}

Fig.~\ref{fig:pomdp} shows a single time step of interaction in a POMDP.
All signals inside the dashed box are fully observed by the agent, while information regarding the state is only obtained via observations.
The dashed line does not necessarily represent boundaries of physical embodiment: the state may refer to, e.g., the acting robot's joint angles.
In a finite-horizon POMDP, the task ends after a specified number of time steps.

{Formally, }a finite-horizon POMDP is a tuple $\langle h, S, A, \Omega, T, O, R,
\gamma, b_0\rangle$ where $h\in\mathbb{N}$ is the time horizon of the
problem, $S$, $A$, and $\Omega$ are the state, action, and observation sets, respectively,
$T$ is the
stochastic state transition model, such that {$T(s', a, s) \coloneqq P(s' \mid s, a)$} is the probability that the next state\footnote{We indicate quantities on the next time step by symbols such as $s'$, and on the current time step by plain symbols such as $s$.} is $s'$ given that the current state is $s$ and action $a$ is executed,
$O$ is the
probabilistic observation model, such that {$O(o',s',a) \coloneqq P(o' \mid s', a)$} is the probability that observation $o'$ is perceived if state $s'$ was reached after executing action $a$ on the previous time step,
$R$ is a bounded reward function, such that $R(s,a)$ is the real-valued reward obtained executing action $a$ in state $s$, $0 \leq \gamma \leq 1$ is a discount factor that determines the relative value of immediate and future {reward}\footnote{The discount factor is not strictly necessary in the finite-horizon case, but we include it for generality and consistency with the infinite-horizon case.}, and $b_0$ is an initial {probability mass function (pmf) over states} such that $b_0(s)$ is equal to the probability that the initial state of the system is $s$.

{The state may include parts relating to an external environmental state and the robot system's internal state.
Partial observability can exist in both of these parts.
To manage the partial observability, the agent} maintains {an estimate of the system state $b$} by applying Bayesian filtering.
The {estimate is a pmf over the system state, and} is a sufficient statistic of the history of past actions and observations.
As opposed to the history which grows over time, the {estimate} has a fixed representation size making it a desirable alternative.
In the POMDP literature these {estimates} are referred to as \emph{belief states}.
The belief state at any time step is defined as the conditional probability distribution over the state given the history of past actions and observations.
As defined above, the initial belief state before taking any actions or perceiving any observations is $b_0$.
Given the current belief state $b$ and action $a$ and a resulting observation $o'$, the updated belief state $b'$ is obtained by Bayes' rule:
\begin{equation}
    \label{eq:belief_update}
    {b'(s') = \frac{P(o'\mid s',a)\sum\limits_{s} P(s'\mid a,s)b(s)}{\eta(o'\mid b, a)}},
\end{equation}
where the denominator is the prior probability of observing {$o'$, i.e., $\eta(o'\mid b, a) = \sum_{s'}P(o'\mid s',a)\sum_{s}P(s'\mid a,s)b(s)$}.
We shall use a shorthand $\tau$ for the Bayes filter defined via Eq.~\eqref{eq:belief_update} as $b' = \tau(b, a, o')$.

A solution for a finite horizon POMDP is {often represented as} a sequence $\pi=(\pi_1, \pi_2, \ldots, \pi_h)$ of policies, where the policy
$\pi_t$ is a mapping from belief states to actions when $t$
decisions remain until the end of the horizon. 
Executing a policy from belief state $b$ when $t$ decisions remain means to first take the action $a = \pi_{t}(b)$, perceive the observation $o'$, and set $b' = \tau(b,a,o')$. Then, execute the policy from $b'$ when $(t-1)$ decisions remain, {now choosing the action according to $\pi_{t-1}$,} repeating until no decisions remain.
The expected {total discounted reward} collected when executing a policy quantifies its performance.
An optimal policy maximizes the performance, as we will see next.

The optimal value function $V_t^*(b)$ is defined to be equal to the expected {total discounted reward} when the agent executes an optimal policy from belief state $b$ when $t$ decisions remain\footnote{Many optimal policies may exist, but the optimal value function is unique.}.
Clearly, $V_1^*(b) = \max_a R(b,a)$, where we use the
shorthand {$R(b,a) \coloneqq \sum_s R(s,a)b(s)$} for the expected
immediate reward of taking action $a$ in belief state $b$.
We can now recursively define $V_t^*$ applying Bellman's principle of optimality~\cite{bellman1966dynamic}: an optimal action when $t$ decisions remain must maximize the sum of the expected immediate reward and the expected future {total discounted reward} for the remaining $(t-1)$ decisions.
Formally,
\begin{equation}
    \label{eq:finitehorizon_optimal_v}
    V_t^*(b) = \max\limits_{a}\left[R(b,a) + \gamma\sum\limits_{o'}\eta(o'\mid b,a)V_{t-1}^*(b') \right],
\end{equation}
where $b' = \tau(b,a,o')$.
As it is not known which next observation $o'$ will be perceived after taking the current action, we take the expectation over $o'$.
The value function $V_t^\pi$ of an arbitrary policy $\pi$ is characterized by a similar recursion $V_t^\pi(b) = R(b,\pi_t(b)) + \gamma\sum_{o'}\eta(o'\mid b,\pi_t(b))V_{t-1}^{\pi}(\tau(b,\pi_t(b),o'))$, starting from $V_1^\pi(b) = R(b,\pi_1(b))$.
Two policies can be compared in terms of their value functions.
The optimal value function satisfies $V_t^*(b) \geq V_t^{\pi}(b)$ for all $b, t$, and $\pi$.

To characterize optimal policies, consider the action-value function $Q_t^*(b,a)$ defined as the expected {total discounted reward} when action $a$ is executed in belief state $b$ when $t$ {decisions} remain, and an optimal policy is executed for the remaining $(t-1)$ decisions thereafter.
We have $Q_1^*(b,a) = R(b,a)$, and for $t\geq 2$,
\begin{equation}
    \label{eq:finitehorizon_q}
    Q_t^*(b,a) = R(b,a) + \gamma\sum\limits_{o'}\eta(o'\mid b,a)V_{t-1}^*(b'),
\end{equation}
where $b' = \tau(b,a,o')$.
The action-value function $Q_t^\pi$ of an arbitrary policy $\pi$ is obtained {by} replacing the value function on the right hand side of Eq.~\eqref{eq:finitehorizon_q} by $V_{t-1}^\pi$.
An optimal action is determined by $\pi_t^*(b) \in \argmax_{a} Q_t^*(b,a)$, which maximizes the right hand side of Eq.~\eqref{eq:finitehorizon_optimal_v}.

\subsection{Infinite horizon}
\label{subsec:infinite_horizon}
{Unending tasks can be modeled using} infinite horizon POMDPs.
{We now briefly review their key concepts.}
Technical details and proofs are found, e.g., in~\cite{hauskrecht2000value,puterman1994markov}.

As the horizon $h$ tends towards infinity, the discount factor is constrained to $0 \leq \gamma < 1$ to ensure the {total discounted reward} is well-defined.
In the infinite horizon case there is always an infinite number of decisions remaining until the end of the horizon. Informally, this is the reason why a solution of an infinite horizon POMDP is a so-called stationary policy, a single policy
$\pi$ applied on every time step.

The
infinite-horizon optimal value function $V^*$ is the
unique fixed point of Eq.~\eqref{eq:finitehorizon_optimal_v} when
$t\to\infty$.
Formally, $\lim_{t\to\infty}\sup_b |V_t^*(b)
- V^*(b)| = 0$.  
Note the
lack of subscripts in $V^*$, indicating its infinite horizon nature.  
The infinite horizon optimal action-value function $Q^*(b,a)$ is defined as in Eq.~\eqref{eq:finitehorizon_q}, replacing the value function on the right hand side by $V^*$.  
An optimal policy is defined via $\pi^*(b) \in \argmax_a Q^*(b,a)$.  

The finite horizon optimal value function $V_t^*$ approximates $V^*$ with a bounded error.
In particular,
let $\epsilon = \sup_b |V_{t}^*(b) - V_{t-1}^*(b) |$ be the least upper bound of the change between consecutive iterations of Eq.~\eqref{eq:finitehorizon_optimal_v}.  Then, $\sup_b |V_{t}^*(b) - V^*(b) |
\leq \gamma\epsilon/(1-\gamma)$.  That is, if we wish to approximate
$V^*$ with precision $\delta$, we can take any $V_{t}^*$  for which $\epsilon$ is less than $\delta(1-\gamma)/\gamma$.
{Solution algorithms for infinite horizon POMDPs can exploit this fact.} 
By solving $V_t^*$ for a sufficiently large $t$ and returning an optimal policy $\pi_t^*$ {we obtain} an approximation with an error at most $\delta$.

\section{POMDP  algorithms}
\label{sec:solvers}
POMDP solution algorithms compute optimal or {approximately optimal} value
functions and policies.  We review the two main categories of solution
algorithms: offline and online algorithms.  We close the section by
discussing heuristics that are widely used for computing POMDP
policies in {robotic applications}. 
We use the following
classical POMDP problem as a running example.
\begin{example*}[Tiger~\cite{cassandra1994acting}]
An agent stands in front of two doors, one on the right and one on the left. 
A tiger is behind one door and a treasure behind another, but
the agent does not know a priori behind which door the tiger is. The
agent may open either door, gaining {$10$ reward} for finding the
treasure or $-100$ for finding the tiger. Alternatively the agent may
listen, gaining $-1$ reward. When listening, the agent makes the
correct observation about the tiger's location with probability
$0.85$. The problem resets to a uniformly random tiger location after
opening either door.
\end{example*}
We have chosen the Tiger problem  as a didactic example for its simplicity. While it lacks meaningful dynamics, which is common in robot tasks, it clearly illustrates the  principle used in the  POMDP framework for decision making under uncertainty.
Another example  with more realistic robot dynamics is the \emph{Tag} problem, in which an agent's goal is to search for and tag a moving opponent~\cite{pineau2003point}.
{Specifically, in contrast to the Tiger problem, the \emph{Tag} problem includes the robot and opponent location as parts of the state that may change as a result of an action while also affecting the reward and the observation.}

\subsection{Offline algorithms}
\label{subsec:offline_algorithms}
The objective of offline algorithms is to compute a policy for all
possible belief states before starting policy execution.  At execution time,
an action is simply looked up from the policy computed
offline.  Offline algorithms are usually based on {iteratively applying} the Bellman optimality principle~\cite{astrom1965optimal,bellman1966dynamic}. Solving a
POMDP for horizon $h$ is recursively broken down into solving
subproblems for horizons $h-1, h-2, \ldots, 1$, similar to the
derivation of Eq.~\eqref{eq:finitehorizon_optimal_v}.  We review basic
principles of exact and approximate offline algorithms, and refer to
the
surveys~\cite{monahan1982state,white1991survey,lovejoy1991survey,hauskrecht2000value,shani2013survey}
for further details.

\begin{figure}[t]
    \centering
\includegraphics[width=0.5\columnwidth]{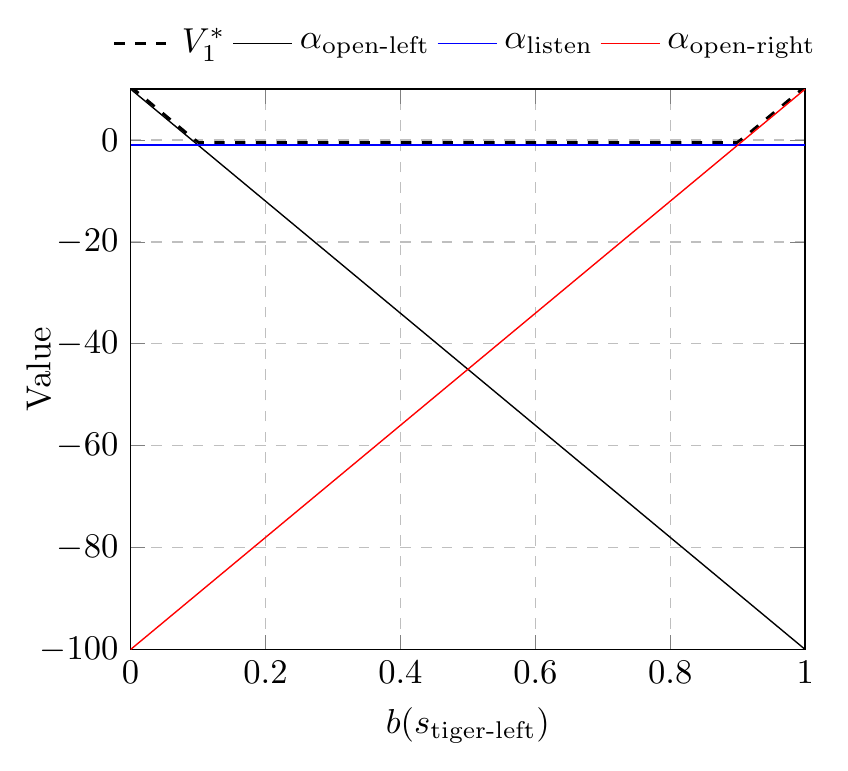}~\includegraphics[width=0.5\columnwidth]{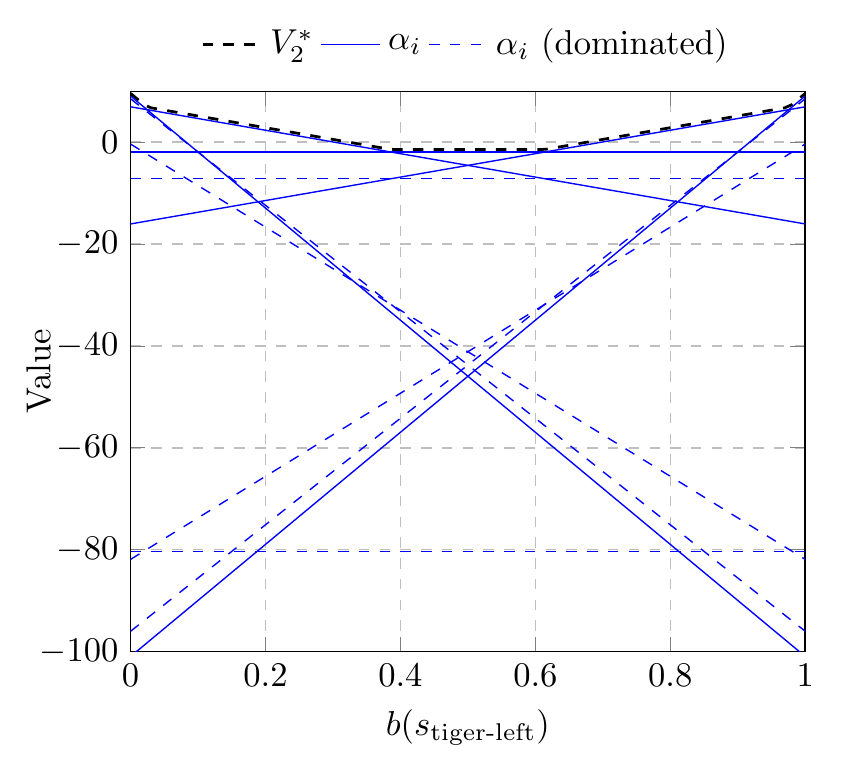}
    \caption{Optimal value functions (dashed black lines) and their $\alpha$-vectors (solid lines) in the Tiger problem. \textbf{Left:} the one-step optimal value function
      $V_1^*$. \textbf{Right:} the two-step optimal value function $V_2^*$. Only $\alpha$-vectors drawn with solid lines are needed to represent $V_2^*$. Dominated $\alpha$-vectors indicated by dashed blue lines do not affect $V_2^*$ and can be pruned.}
  \label{fig:alphavectors}
\end{figure}

The key property employed by offline POMDP algorithms is that the
horizon-$t$ optimal value function $V_t^*$ is a piecewise linear and
convex (PWLC) function of the belief
state~\cite{sondik1971optimal,smallwood1973optimal}.  Formally,
$V_t^*$ is the convex hull of a finite set of
$|S|$-dimensional hyperplanes, also known as $\alpha$-vectors:
\begin{equation}
\label{eq:optimal_value_t_alpha_vectors}
    V_t^*(b) = \max\limits_{\alpha \in \Gamma_t} \sum\limits_{s}\alpha(s)b(s),
\end{equation}
where $\Gamma_t$ is the set of $\alpha$-vectors that represents the
horizon-$t$ optimal value function.  It is easy to see that $\Gamma_1
= \{ \alpha^{a} | a\in A: \alpha^{a}(s) = r(s,a)\}$ since only the
expected immediate reward matters.  {We may prove} by induction
that $V_t^*$ for every finite $t$ is also PWLC.
Fig.~\ref{fig:alphavectors} (left) shows the optimal one-step value
function for the Tiger problem.  In the tiger problem
$b(s_\text{tiger-left}) = 1 - b(s_\text{tiger-right})$, so the belief
{is} represented by a single real number corresponding to the
probability that the tiger is behind the left door.  As seen from the
figure, the $\alpha$-vectors corresponding to each action dominate
in different regions of the belief space (the horizontal axis).  For
instance, if the agent is certain that the tiger is on the left,
opening the right door is optimal.

Given $\Gamma_{t-1}$ as input, exact offline algorithms iteratively calculate the set $\Gamma_{t}$
of $\alpha$-vectors representing $V_{t}^*$ for any number of time steps $t$ remaining. 
The solution for horizon $(t-1)$ is used as a building block of the solution for horizon $t$ by applying the principle of dynamic programming.
{We show an example of a single iteration 
in Fig.~\ref{fig:alphavectors}, where we use the set $\Gamma_1$ illustrated
on the left together with the POMDP model to compute the set
$\Gamma_2$ illustrated on the right.}  By keeping track of the
immediate actions associated with each $\alpha$-vector, {we can recover optimal
policies $\pi_t^*$ from $\Gamma_t$.}

Exact offline algorithms are a suitable choice for small problems with up to a few dozen states.
The complexity of the value function representation depends on the number of actions and observations.
In the worst case,
the number of $\alpha$-vectors required is $|\Gamma_t| = |A||\Gamma_{t-1}|^{|\Omega|}$, which grows rapidly as {a} function of $t$.
However, {not} all of the $\alpha$-vectors may be necessary to represent
the value function exactly.  An $\alpha$-vector is \emph{dominated} if it is not optimal for any belief state.
A dominated $\alpha$-vector can
be removed from $\Gamma_t$ without affecting the value function
representation.
{Fig.~\ref{fig:alphavectors} (right) shows examples of dominated $\alpha$-vectors by the dashed lines.}
The step of
identifying and removing dominated $\alpha$-vectors is known as
\emph{pruning}, see,
e.g.,~\cite{monahan1982state,white1991survey,lovejoy1991survey,hauskrecht2000value}.
Even with pruning, the growth in the number of $\alpha$-vectors makes exact offline methods infeasible for problems with large action and observation spaces.

Point-based value iteration
algorithms~\cite{shani2013survey,pineau2006anytime,smith2004heuristic,spaan2005perseus,kurniawati2008sarsop}
are a class of approximate offline algorithms that represent the value
function by maintaining one $\alpha$-vector per each belief state in a
finite set $B$ of belief states.  
The total number of
$\alpha$-vectors required to represent a value function is bounded by
$|B|$, which leads to computational savings compared to exact
algorithms.  
Point-based methods are a useful choice in robotic tasks such as grasping~\cite{hsiao2007grasping} where the set of belief states that are \emph{reachable} under a reasonable policy is only a small subset of the full belief space.

Point-based algorithms may
select belief states to be included in $B$, e.g., randomly~\cite{spaan2005perseus}, by minimizing error
bounds~\cite{smith2004heuristic,pineau2006anytime}, or to {better} approximate the set of reachable belief states~\cite{kurniawati2008sarsop}.
Due to the PWLC property of the value function, {we may still easily recover} the value
function at an arbitrary belief state from a
point based representation~\cite{pineau2006anytime}. {We update the value
function representation by an iterative procedure similar
to the exact case, but only consider the belief states in $B$.}
While iterating, {we may add new beliefs to $B$, and bound the
approximation quality dependent on how densely $B$ covers the entire space of possible beliefs~\cite{pineau2006anytime}.}
Point-based algorithms can address significantly larger problems than exact offline algorithms {by sacrificing optimality}~\cite{pineau2006anytime}.
{However, for very large problems, even point-based algorithms are not computationally tractable.
For such problems, we can apply online algorithms.}

\subsection{Online algorithms}
\label{subsec:online_algorithms}

Online algorithms interleave policy computation and
execution. The algorithms compute an optimal or {approximately} optimal
action for the current belief state $b_0$, execute the action, make an
observation, and continue by computing an action for the resulting new
belief state. 
Since a solution is {only} computed for the current belief state $b_0$, online algorithms can address larger problems than offline algorithms.
Online algorithms commonly work by searching over reachable belief states or equivalent
action-observation sequences.
We review forward search in online algorithms, and discuss modern sampling-based implementations that have
led to major improvements in the scalability of POMDP algorithms to
large problem domains. For a general survey of online
algorithms, we refer the reader to~\cite{ross2008online}.

\begin{figure}[t]
    \centering
    \includegraphics[width=\columnwidth]{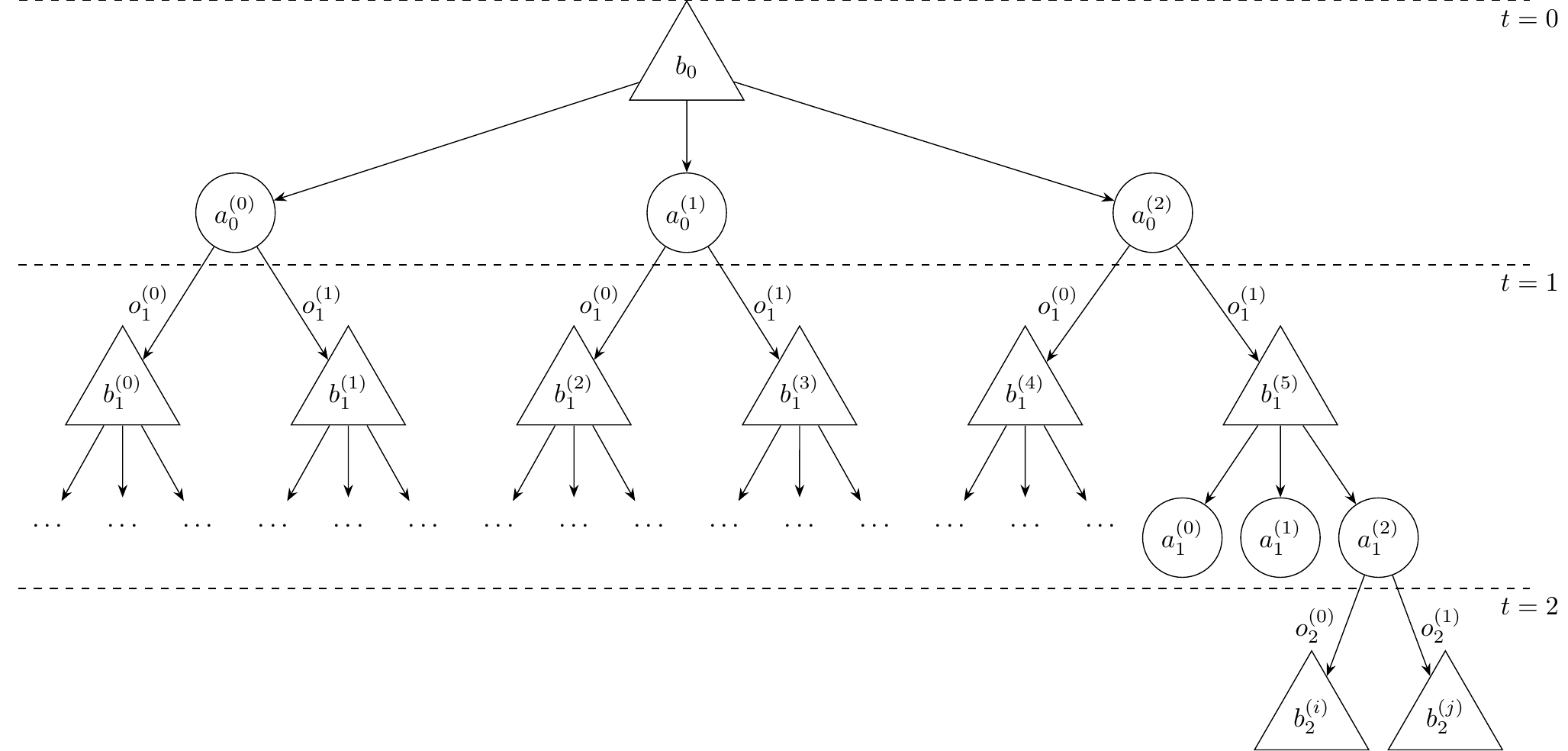}
    \caption{A tree over action-observation sequences for the
      Tiger problem. At triangular decision nodes on time step $t$, the agent selects an
      action $a_t^{(0)}$ (open-left), $a_t^{(1)}$ (open-right), or $a_t^{(2)}$
      (listen).
      At circular chance nodes, the agent receives an
      observation $o_t^{(0)}$ (hear-left) or $o_t^{(1)}$ (hear-right).
      Time steps are indicated by the
      section labels on the right hand side. Note that the time steps here denote steps since start of the task, \emph{not} steps remaining.
      The tree is shown
      fully expanded until the observation at $t=1$, and partially
      thereafter.
      In node labels, subscripts denote time step and superscripts in parentheses an index.}
    \label{fig:online_tree}
\end{figure}

Given a fixed initial belief state $b_0$, let us consider all possible
action-observation sequences $(a_0,o_1,\ldots,a_{t-1},o_t)$ of
length $0 \leq t \leq h$.
Such sequences are commonly referred to as \emph{histories}.
As
shown in Fig.~\ref{fig:online_tree}, we can build a tree over histories with the root node corresponding to $b_0$.
Each triangular decision node corresponds to the history obtained by
concatenating actions and observations along the path
from the root to the node.  
The related belief states may be computed by
repeatedly applying the Bayes filter $\tau$.

Online algorithms build a search tree over histories, and return an
optimal action to take at the initial belief
state~\cite{ross2008online}.  In practice, {we use} the
tree structure to compute the optimal action-value
function $Q_h^*(b_0,a)$.  For example, {consider calculating the optimal action-value of listening on the first time step in the Tiger problem.}
Contrasting
Eq.~\eqref{eq:finitehorizon_q} with Fig.~\ref{fig:online_tree} we
see that the optimal action-value $Q_h^*(b_0, a_0^{(2)})$ at the root node
depends on the expected immediate reward $R(b_0,a_0^{(2)})$ and
$V_{h-1}^*(b')$, where $b'$ is either one of the successor beliefs
$b_1^{(4)}$ or $b_1^{(5)}$ reached after taking the listening action $a_0^{(2)}$ and perceiving $o_1^{(0)}$ (hear-left) or $o_1^{(1)}$ (hear-right), respectively.
We have $Q_h^*(b_0,a_0^{(2)}) = R(b_0,a_0^{(2)}) + \gamma\sum_{o'}\eta(o'\mid b_0, a_0^{(2)})V_{h-1}^*(b')$ with $b'$ as above depending on $o'$.
To proceed, we recall $V_{h-1}^*(b') = \max_a Q_{h-1}(b',a)$, which is computed by expanding the next layer of chance nodes under $b'$.
This recursion terminates at the leaves of the tree.
A naive online algorithm builds a complete search tree as explained
above, computing belief states, expected {reward}, and observation
probabilities to return an optimal
action $\argmax_a Q_h^*(b_0,a)$.  Infinite horizon problems are
tackled by searching over a finite horizon $h$ that reaches the
desired approximation quality as described in
Sect.~\ref{subsec:infinite_horizon}.

{The number of nodes in the $(t+1)$th level of the search tree is equal to the number of nodes in the $t$th level multiplied by the branching factor $|A||\Omega|$. Therefore, the }
$t$th level of the tree has $|A|^t|\Omega|^t$ decision nodes,
corresponding to the number of histories of
length $t$.
This exponential increase of tree size as {a} function of its
depth is a major challenge for online algorithms based on tree search.
Several techniques have been proposed to address this
challenge~\cite{ross2008online}.  For example, branch-and-bound
pruning {traverses actions in a descending order according to an upper bound and}
avoids expanding tree branches that are known to be
suboptimal by using lower and upper bounds for the optimal
action-value function.  {The expansion order of branches is sometimes prioritized based on
heuristics.}  Expanding one branch of the tree may reveal
that another branch is suboptimal and does not need to be expanded.

The need to explicitly compute the belief states associated with each node is also prohibitively expensive in problems with large state spaces. 
Instead of explicit belief state computation, state-of-the-art online algorithms for POMDPs draw state samples from the
initial belief state and propagate them through the nodes of the
search tree by drawing samples from the dynamics model and
observation model to determine successor nodes.
{A Monte Carlo procedure using these samples is used to compute estimates of
action-values.}

In sampling-based online algorithms for POMDPs, e.g.,~\cite{silver2010monte,somani2013despot,sunberg2018online}, belief states are represented as collections of particles at decision nodes, and a
simulator that allows sampling of {the next state, reward, and
observation} is used for constructing the tree and estimating the
action-value functions.
It is sufficient to be able draw samples from the state transition and observation models, as no explicit belief state tracking by Bayes filtering is necessary during planning.
As a simulator is sufficient for planning, these methods are very flexible and can also handle {robotic tasks} where an explicit model may be unavailable.
The algorithms provide asymptotic optimality guarantees under appropriate assumptions: as the number of iterations tends towards infinity, the solution will approach an optimal solution.

\begin{figure}[t]
    \centering
    \includegraphics[width=\columnwidth]{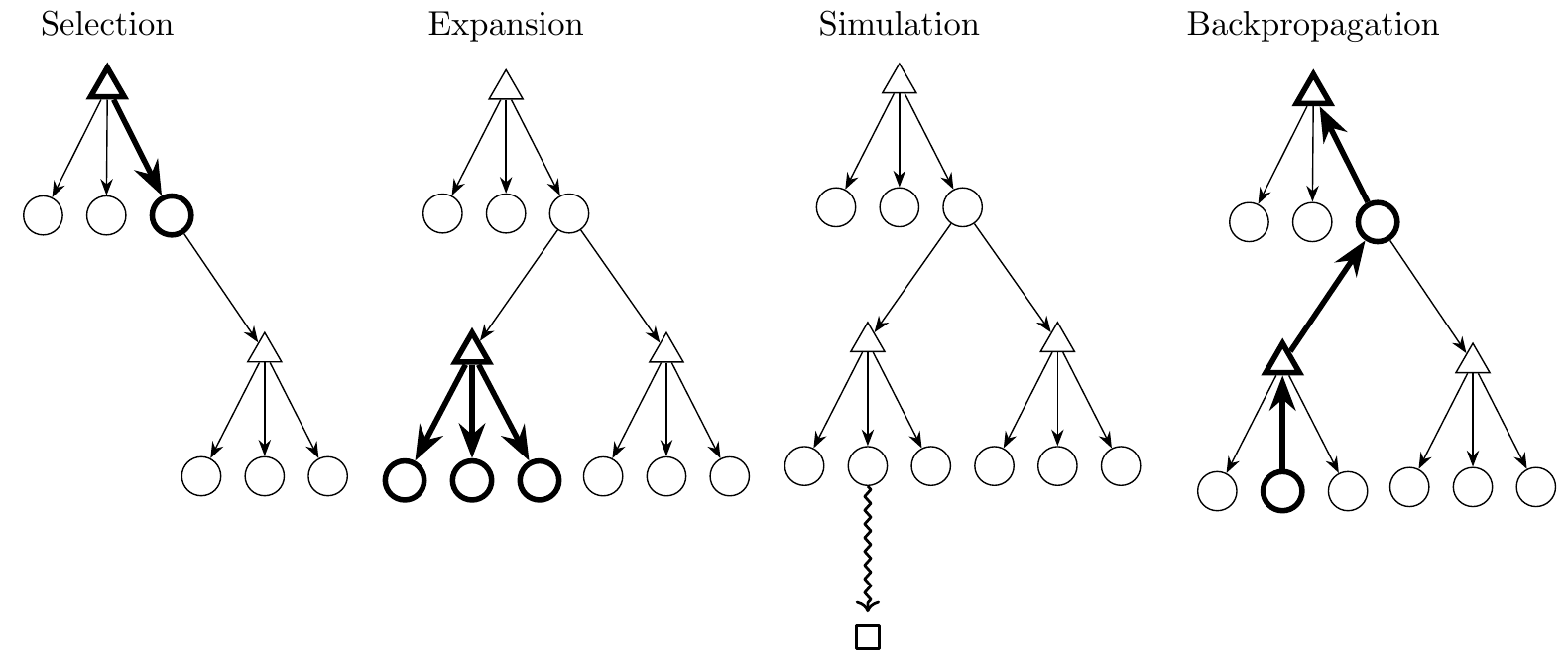}
    \caption{POMCP repeats the four phases (from left to right) of
      selection, tree expansion, simulation/rollout, and
      backpropagation to build a sparse search tree over
      action-observation histories. {Triangle/circle node shapes are as
      in Fig.~\ref{fig:online_tree}. Annotations are omitted for
      clarity.}}
    \label{fig:pomcpmcts}
\end{figure}

Sampling-based online algorithms differ in their strategies for constructing the search tree.
POMCP~\cite{silver2010monte} relies on Monte Carlo tree search (MCTS; e.g.,~\cite{browne2012survey}) to iteratively build the search tree as illustrated in Fig.~\ref{fig:pomcpmcts}.
In the selection phase, {we traverse the current search tree by selecting actions using the upper confidence bound (UCB)
  multi-armed bandit rule that balances between exploration and
  exploitation.}
Once {we reach a leaf node, we add new nodes} to the tree in the expansion phase.
In the simulation phase, {we perform a simulation \emph{rollout}} to estimate the long-term value of the leaf node that was reached.
In the backpropagation phase, {we use} the information from the rollout to update the action-value estimates of nodes along the path from the root node to the leaf node.
POMCPOW~{\cite{sunberg2018online}} also uses MCTS, with additional techniques applied for dealing with continuous action and observation spaces.
In contrast, determinized sparse partially observable trees
(DESPOT)~\cite{somani2013despot} constrains {the} search to a finite set
of randomly sampled \emph{scenarios}, {and builds} a search tree that covers
$|A|^tK$ nodes at depth $t$, where $K$ is the number of scenarios. The random scenarios determine which branches of the search tree will be expanded.

Online algorithms may also search for a solution in a limited set, such as finite state controllers~\cite{bai2010monte,bai2014integrated,pajarinen2017robotic} or other parametric representations~\cite{martinez2009bayesian,buffet2009factored}.
These {so-called policy search algorithms gain scalability by limiting the search space, but often lose the ability to bound solution quality.}

We conclude by contrasting the reviewed online forward search algorithms to offline dynamic programming based solutions from the previous subsection.
Early online algorithms~\cite{ross2008online} that build a complete search tree or use admissible heuristics to prune it with explicit computation of belief states will return an optimal solution to finite-horizon problems, {similarly to} exact offline algorithms.
MCTS-based algorithms~\cite{silver2010monte,sunberg2018online} use a greedy heuristic in the selection phase to choose how to expand the search tree, and aggressively exploit {the} most promising branches of the search tree.
They can produce search trees that are shallow in some parts and deep in other parts,  potentially discovering solutions that require a very long planning horizon.
On the other hand, they are also prone to getting ``trapped'' in deep branches of the search tree preferred by the heuristic but ultimately not useful, see, e.g.,~\cite[Sect.~2.3.]{Munos14}.
While the asymptotic optimality property guarantees the search will eventually avoid such traps, the number of samples required for this may be infeasible in practice.
Online algorithms using scenarios~\cite{somani2013despot} can avoid getting stuck due to poor choice of heuristics, but do not exploit promising branches of the search tree as aggressively.
The scenario-based methods {are therefore in the middle} of the spectrum ranging from greedy MCTS solutions to conservative online algorithms {that are} more similar to offline dynamic programming methods.

\subsection{Heuristics}
\label{subsec:bounds_heuristics}
Heuristics are computationally inexpensive ways to obtain a solution to a POMDP based on simple principles such as taking an action that yields the greatest expected immediate reward.
Using heuristics comes at the cost of losing the approximation
guarantees provided by offline and online algorithms reviewed above.
We aim to provide the reader with a basic understanding of the trade-offs and assumptions underlying the most widely applied heuristics in the literature to find solutions to robotic POMDP problems.
For a more comprehensive overview of heuristics, we refer the reader
to~\cite{hauskrecht2000value,bertsekas2005dynamic}.

The greedy policy heuristic takes an action that maximizes the expected immediate reward: $\argmax_{a} R(b,a)$.
It is quick to compute from the current belief state and reward function.
However, it ignores any reasoning about long-term effects of actions.
{There is also no guarantee on how much worse the greedy policy} is compared to an optimal solution.

The QMDP heuristic~\cite{littman1995learning} first computes the
optimal action-value function $Q_{\text{MDP}}^*\colon S\times
A\to\mathbb{R}$ of the underlying \emph{fully observable} Markov
decision process (MDP)~\cite{puterman1994markov} where the agent
always perceives the true state.  QMDP then computes an upper bound
\begin{equation}
\label{eq:qmdp}
    \hat{V}(b) \coloneqq \max\limits_{a}\sum\limits_{s}b(s)Q_{\text{MDP}}^*(s,a)
\end{equation}
for the optimal value function $V^*$ of the POMDP.  The QMDP control
policy is then obtained by taking an action that maximizes the above
equation.  Eq.~\eqref{eq:qmdp} defines $\hat{V}(b)$ as the best
expected value in the MDP, when uncertainty about the state will
disappear after the current action.  Therefore, QMDP is not well
suited for problems where explicit information gathering is required
for success, but is computationally much cheaper than a full POMDP
solution.

Open loop feedback control (OLFC)~\cite{bertsekas2005dynamic}, also known as receding horizon control (RHC) or
model predictive control (MPC), is a well-known method from
stochastic control that can also be applied to finite-horizon
POMDPs.  OLFC computes an optimal
\emph{open loop policy}, i.e., a sequence of actions with the greatest
expected return for the current belief state.  Then, the first action
in the sequence is executed, and an observation is perceived.  A
posterior belief state is computed using the Bayes filter, and the
process is repeated.  This interleaving of computation of open loop
policies and observing feedback in the form of observations gives the
method its name.  There are many fewer possible action sequences than
closed loop policies, making the problem computationally cheaper.  The
performance of OLFC is at least as good as that of simply executing
the entire optimal open loop action
sequence~\cite{bertsekas2005dynamic}, but no error bound compared to
an optimal solution is known.

\section{POMDPs in Robotics}
\label{sec:applications}
This section discusses robotic application domains where POMDPs have
been heavily used: localization and navigation, autonomous driving, search and tracking, manipulation, human-robot interaction, and multi-robot coordination. 
Each application domain has its specific
challenges related to partial observability. How the POMDP model is
specified depends not only on the application domain but often also on
the specific application requirements. In addition, a large variety of
different POMDP algorithms have been used.

\begin{table*}[!t]
\centering
\caption{Key sources of uncertainty and common POMDP solution algorithms by application area in the literature. 
The bottom row collects other less frequently applied algorithms by application.
Acronyms used include Cross Entropy (CE), Monte Carlo (MC), Variational Bayes (VB), Value Iteration (VI), 
Policy Gradient (PG), Policy Iteration (PI), Policy Search (PS).}
\label{tab:algorithms_applications}
\begin{tabularx}{\textwidth}{LLLLLLL}
\toprule
 & 
 \textbf{Localization \& Navigation} & 
 \textbf{Autonomous Driving} & 
 \textbf{Search \& Tracking} & 
 \textbf{Manipulation} & 
 \textbf{Human-Robot Interaction} & 
 \textbf{Multi-Robot Coordination} \\
\midrule
 \textbf{Key Sources of Uncertainty} & 
 \textit{State}\newline robot location, environment map & 
 \textit{State}\newline locations of traffic agents\newline
 \textit{Dynamics}\newline control effects, others' behaviors
 \textit{Perception}\newline poor weather, occlusion &
 \textit{State}\newline target location \newline
 \textit{Dynamics}\newline target behavior & 
 \textit{State}\newline object pose \newline
 \textit{Dynamics}\newline grasp success/failure \newline
 \textit{Perception}\newline visual ambiguity, occlusion & 
 \textit{State}\newline human intention &
 \textit{State}\newline information private to each robot\\
\midrule
 \textbf{Point-based} \\ %
    \hspace{0.5em}SARSOP~\cite{kurniawati2008sarsop} &
    & 
    & 
    \cite{gupta2017decision,atanasov2014nonmyopic} & 
    \cite{monso2012pomdp,horowitz2013interactive,koval2016pre} & 
    \cite{schmidt2010learning,chen2018planning,petric2019hierarchical} & 
    \\ %
    \hspace{0.5em}PBVI~\cite{pineau2006anytime} &
    & 
    & 
    & 
    \cite{hsiao2007grasping} & 
    \cite{atrash2009bayesian,atrash2010bayesian} & 
    \\ %
    \hspace{0.5em}HSVI~\cite{smith2005point} &
    & 
    & 
    \cite{sridharan2010planning} & 
    \cite{broz2011designing} & 
    & 
    \\ %
    \hspace{0.5em}Perseus~\cite{spaan2005perseus} &
    & 
    & 
    & 
    & 
    \cite{rosen2020mixed} & 
    \cite{spaan2008cooperative}
    \\
    \hspace{0.5em}S-Perseus~\cite{poupart2005exploiting} &
    & 
    & 
    \cite{capitan2013decentralized,spaan2010active} & 
    & 
    \cite{hoey2010automated,taha2011pomdp} & 
    \cite{capitan2013decentralized,spaan2010active} \\
\midrule
 \textbf{Tree search} \\ %
    \hspace{0.5em}POMCP~\cite{silver2010monte}\xdef\tempwidth{\the\linewidth} &
    \cite{lauri2016planning}{,\cite{kim2021plgrim}} & 
    \cite{sunberg2017value} & 
    \multicolumn{1}{>{\raggedright\arraybackslash}m{\tempwidth}}{\cite{wandzel2019multi,xiao2019online,wang2020pomp,goldhoorn2018searching,zheng2021}} & 
    \cite{xiao2019online} & 
    & 
    \cite{goldhoorn2018searching} \\
    \hspace{0.5em}DESPOT~\cite{somani2013despot}\xdef\tempwidth{\the\linewidth} &
    \cite{lee2021} & 
    \multicolumn{1}{>{\raggedright\arraybackslash}m{\tempwidth}}{\cite{bai2015intention}, {\cite{sunberg2018online}}, \cite{pusse2019hybrid,luo2018autonomous,cai2020hyp,ychsu2020}}  & 
    & 
    \cite{li2016act,garg2019learning} & 
    \cite{unhelkar2019semi,unhelkar2020decision}& 
    \\
    \hspace{0.5em}ABT~\cite{kurniawati2013online,klimenko2014tapir}  &
    & 
    \cite{hubmann2017decision} & 
    \cite{vanegas2016uav} & 
    & 
    & 
    \\
\midrule
 \textbf{Policy search} \\ %
    \hspace{0.5em}MCVI~\cite{bai2010monte}  &
    \cite{grady2015extending} & 
    & 
    \cite{bai2012unmanned} & 
    & 
    & 
    \\
    \hspace{0.5em}PPGI~\cite{pajarinen2017robotic}  &
    & 
    & 
    & 
    \cite{pajarinen2014robotic,pajarinen2017robotic,pajarinen2020pomdp} & 
    \cite{hoelscher2018utilizing} & 
    \\
    \hspace{0.5em}Random shooting  &
    \cite{dallaire2009bayesian,martinez2009bayesian} & 
    & 
    & 
    & 
    & 
    \cite{wang2017anticipatory}\\
\midrule
 \textbf{Heuristic} \\ %
    \hspace{0.5em}QMDP~\cite{littman1995learning} &
    \cite{koenig1998xavier}{,\cite{kim2021plgrim}} & 
    \cite{wei2011point} & 
    & 
    & 
    & 
    \cite{spaan2008cooperative}\\
    \hspace{0.5em}Greedy &
    & 
    & 
    \cite{eidenberger2010active,valimaki2016optimizing} & 
    & 
    & 
    \cite{ahmadi2019safe}\\
\midrule
 \textbf{Others} &
    \textit{Pre-image back-chaining}~\cite{kaelbling2017pre,kaelbling2013integrated}\newline
    \textit{FIRM~}\cite{agha2014firm,agha2014health}\newline
    \textit{MEDUSA~}\cite{jaulmes2007formal}\newline
    {\textit{PLGRIM~}\cite{kim2021plgrim}}\newline
    \textit{rAMDL~}\cite{kopitkov2017no}\newline
    \textit{BPS~}\cite{wang2018bounded}\newline
    \textit{GBS}~\cite{indelman2015planning}
    &
    \textit{Continuous MCVI~}\cite{brechtel2013solving,brechtel2014probabilistic}\newline
    {\textit{POMCPOW~}\cite{sunberg2018online}}\newline
    \textit{MC on predefined policies~}\cite{song2016intention}\newline
    \textit{Tree search~}\cite{ulbrich2015towards}\newline
    \textit{MPDM~}\cite{cunningham2015mpdm}
    & 
    \textit{PG~}\cite{buffet2009factored,zhang2013active}\newline
    \textit{MAA*~}\cite{szer2005maastar,lauri2017multi}\newline
    \textit{Greedy PBVI~}\cite{satsangi2018exploiting}\newline
    \textit{Bayesian Opt.~}\cite{martinez2009bayesian}\newline
    \textit{Nominal belief~}\cite{miller2009pomdp}\newline
    \textit{Search~tree~VI~}\cite{yi2019indoor}\newline
    \textit{VB-PBVI~}\cite{burks2019optimal}\newline
    \textit{Unknown~}\cite{foka2003predictive}
    & 
    \textit{Pre-image back-chaining}~\cite{kaelbling2017pre,kaelbling2013integrated}\newline
    \textit{Most likely state~}\cite{hsiao2008robust}\newline
    \textit{POMDP as set of MDPs~}\cite{chen2016pomdp}\newline
    \textit{Offline tree policy~}\cite{zhou2017probabilistic}
    & 
    \textit{Oracular POMDP}~\cite{armstrong2007oracular,rosenthal2011modeling}\newline
    \textit{LSPI~}\cite{lagoudakis2003least,wang2017anticipatory}\newline
    \textit{Exact~}\cite{pineau2003towards}
    & 
    \textit{Nominal belief~}\cite{miller2009pomdp,ragi2014decentralized}\newline
    \textit{Tabular VI~}\cite{bellman1966dynamic,matignon2012coordinated}\newline
    \textit{Dec-MCTS~}\cite{best2019dec,sukkar2019multi}\newline
    \textit{PG~}\cite{buffet2009factored,zhang2013active}\newline
    \textit{MAA*~}\cite{szer2005maastar,lauri2017multi}\newline
    \textit{Masked MC, Graph~CE~}\cite{omidshafiei2017decentralized}\newline
    \textit{Search~tree~VI~}\cite{yi2019indoor}\newline
    \textit{Heuristic PS~}\cite{amato2016policy}\newline
    \textit{Approx.~PI~}\cite{bhattacharya2020multiagent}\newline
    \textit{DSGA~}\cite{corah2019distributed}
    \\
\bottomrule
\end{tabularx}
\end{table*}

Table~\ref{tab:algorithms_applications} summarizes the most commonly used categories of algorithms in robotic applications: point-based, tree search, policy search, and heuristic.
For each algorithm, we indicate application areas where it has been used.
We also highlight key sources of uncertainty (state, dynamics, or perception) in each application area that {POMDP planning techniques address.}
{The last row of the table lists other less frequently used algorithms per application.
For each entry on the last row, the first reference introduces the algorithm and/or application, while remaining references indicate applications.}
By necessity, the table only provides a rough overview and omits aspects such as hierarchical control, details on challenges related to a specific application, or techniques for dealing with high-dimensional observations.
{We provide further details} in the following subsections, adding nuance to the general overview.

\subsection{Localization and Navigation}
\label{sec:localization_navigation}
Self-localization and navigation are basic functionalities of mobile robots.
A warehouse robot needs to be aware of its current
location and be able to navigate to the correct place to pick up the
next batch of goods. In mining, machines need to navigate in dark
underground tunnels and cannot rely on a GPS signal. Outdoor mobile
machines need to cope with changing weather and lighting conditions in
addition to environmental occlusions. Localization and navigation in
cluttered, non-stationary environments with partially occluded visibility and noisy
sensors can be challenging.
POMDP modeling is thus a natural choice in localization and navigation.

\paragraph{Key sources of uncertainty and challenges}
In localization and navigation, a key source of uncertainty is the
location of the robot which cannot be directly observed due to noisy
sensors and sensors that observe only parts of the
environment. Another source of uncertainty is the composition of the
environment which may not be known a priori.

Localization and navigation is often split into localization, path
planning, and path following~\cite{koenig1998xavier}. When localizing
the robot with respect to its surroundings, the environment and a map
of it can be either known beforehand or constructed from observations
by simultaneous localization and mapping (SLAM,
e.g.~\cite{durrantwhyte2006slam}). Path planning creates a sequence of
robot configurations in order to gain more information about the
environment or to reach a given goal location while avoiding obstacles
and dangerous areas. Path following executes the planned sequence of
configurations, taking into account robot dynamics and changes in the
environment or task that may occur while traversing the path.

In simple localization and navigation POMDPs, a 2-D point mass
representation can be used.
{In more complex tasks, we also need to consider the position,
orientation, and velocity of the robot.} 
{Mobile
ground robots often plan paths in 2-D in contrast} to
path planning in 3-D space in robot manipulation (see
Section~\ref{sec:manipulationgrasping}). 
Perceptual aliasing~\cite{filliat2003map} is common due to occluding objects or variable terrain, and leads to multi-modal belief
distributions. For POMDP modeling, localization and navigation can be
challenging due to continuous states, actions, and observations of
which the observation space can be large. Moving obstacles, a dynamic
environment, and changing task goals may increase the difficulty of
the task. Due to the high uncertainty in observations, localization and
navigation is a challenge that has inspired many
classical POMDP benchmarks~\cite{littman1995learning}.

\paragraph{Solution methods}

POMDPs {are} used for both high and low level planning in
localization~\cite{koenig1998xavier}, path
planning~\cite{koenig1998xavier}, and path
following~\cite{agha2014health}. 
The environment is often modeled as a set of states including the
position and the heading of the robot, the location of obstacles as
well as the start and goal locations. {It is common to both
discretize large continuous state, observation, and action
spaces, and to use hierarchical models for the long planning
horizons.}

Among the first to propose to use a POMDP
for indoor navigation, {Koenig and Simmons~\cite{koenig1998xavier} automatically generate the POMDP transition and observation probability distributions from 
a topological map of the environment and prior observation and dynamic models.}
To discretize the problem, {they map the robot pose to a
non-uniform grid, use a discrete set of actions, and map observations to} semantic information such as ``door open'', ``wall'',
or ``door closed''. {They update the POMDP dynamics and observation
models} using expectation-maximization
(EM)~\cite{dempster1977maximum,moon1996expectation} to account for uncertainty in the prior models. EM updates the models of the map, sensors, and actuators using observed
sequences of actions and observations during task execution. For
choosing actions the POMDP model is used at each time step to estimate
the belief over possible pose states using previous sensor readings
and executed actions.
A Gaussian Process~\cite{rasmussen2006gaussian} based POMDP model {for blimp height control
 with continuous states, actions, and observations is learned by~\cite{dallaire2009bayesian} during task execution}.
In~\cite{jaulmes2007formal}, the robot {queries} a human operator to learn the dynamics and observation models.
The queries and when to initiate them are decided by the POMDP policy.
{The complexity of the POMDP model is reduced by limiting the available set of actions in parts of the belief space in~\cite{Grady2013AutomatedMA,grady2015extending}.
This helps to simplify the planning task, for example by reducing the car-like dynamics of a robot to rigid body dynamics.}

\begin{figure}
  \centering
  \includegraphics[width=0.7\columnwidth]{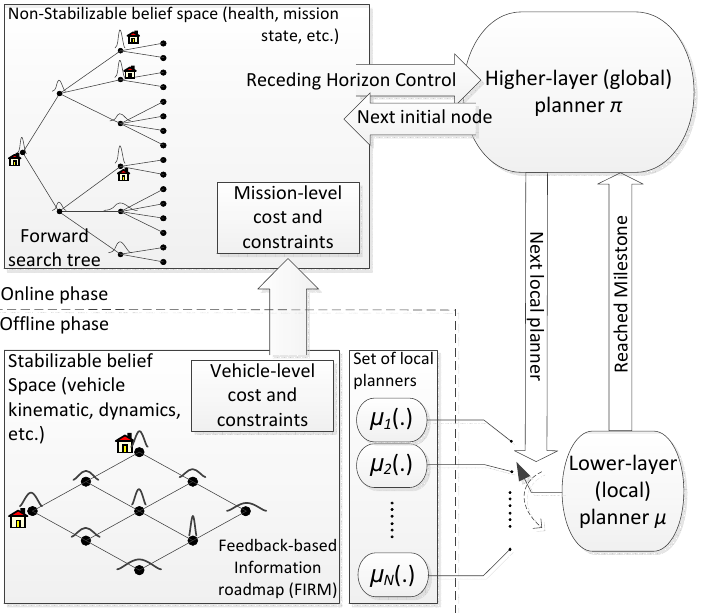}
  \caption{In localization and navigation, hierarchical control is
    typical (Figure from~\cite{agha2014health}). Hierarchical
    control can make planning more efficient, for example, a high
    level policy directs the robot from one waypoint to another and
    the low level policy decides how to travel between the waypoints.}
  \label{fig:localization_navigation}
\end{figure}

{\emph{Hierarchical POMDP models} (e.g., Fig.~\ref{fig:localization_navigation}) are} used for long horizon
planning and complex dynamics found in navigation tasks. 
{The POMDP
is split into high level and low level planning, which makes the planning horizon
for both levels shorter} compared to planning
entirely on the low level. For example, a high level policy {may} direct
the agent from one waypoint to another, and the low level policy
decides how to travel between the waypoints. {Near real-time replanning is possible, since the problem is split into computationally manageable
parts.}
In~\cite{agha2014health} a UAV {delivers} packages while optimizing
mission level statistics such as battery status, or sensor and
actuator health.
{The UAV uses a high-level graph similar to probabilistic roadmaps~\cite{kavraki1996probabilistic} for navigation, and a low-level controller to traverse between the nodes of the graph.}
{A two-layer planner is used in~\cite{indelman2015planning}.
An inner layer predicts posterior beliefs for control actions, and an outer layer uses the predictions to infer a control policy.
No discretization is applied, and solutions are found by MPC.
}
{Temporally extended open loop sequences of actions, called macro-actions, are used in~\cite{lee2021}.
They train a generator to output macro-actions, and show that the temporal abstraction helps to address long planning horizons.
}

Robot navigation can also be framed as a
satisfiability problem. 
The goal is to find a policy that satisfies a set of constraints which guarantee that only belief states considered safe are reached.
A satisfiability
based safe navigation approach is demonstrated by~\cite{wang2018bounded} in simulation with a PR2 robot and
uncertain obstacles.  

{Task and motion planning (TAMP) typically} consists of
localization, navigation, and manipulation (for manipulation, see
Section~\ref{sec:manipulationgrasping}).
{A symbolic pre-image back-chaining~\cite{kaelbling2017pre} based POMDP approach for TAMP is proposed in~\cite{kaelbling2013integrated}.} 
Logical expressions describe sets of belief states, and are used
as goals and subgoals for POMDP planning.
{Interleaving high-level planning for information gathering with execution helps the robot to effectively use any newly gained information.}
{Other recent approaches to TAMP which also address partial observability include~\cite{phiquepal2019,garrett2020}.
These methods may explicitly plan for intelligent information-gathering, which is necessary for many real world tasks.
}

In SLAM, the belief state captures the uncertainty in the robot pose and environment.
\emph{Active SLAM} can employ a belief-dependent reward  function that encourages
information gathering about the state. For example, \cite{martinez2009bayesian} uses a Gaussian Process approximation to model how the
choice of policy parameters influences the uncertainty of the future belief state.
In~\cite{lauri2016planning}, a mobile ground robot explores an environment and estimates an occupancy grid map by choosing short
primitive trajectories as actions. The reward is based on the expected
future information gain about the map. To handle the large discrete
observation space, a variant of POMCP combined with MPC {is} applied.
A factor graph belief representation is used in~\cite{kopitkov2017no}, with a Gaussian approximation in the planning stage.
The objective is to select a trajectory that leads to a posterior belief with the lowest uncertainty.
{New landmarks discovered during exploration are added as new state factors.}
{The belief-dependent reward is a natural way to encourage the robot to take actions that reduce future uncertainty.
However, additional care is needed to avoid getting stuck in locally optimal solutions due to the limited planning horizon.
For example in~\cite{lauri2016planning}, if no sequence of actions yields a high enough information gain, a frontier-based method selects the next exploration target instead of POMDP planning.
}

{The DARPA Subterranean Exploration Challenge is an
  interesting example of large-scale real-world application of
  autonomous localization, mapping and navigation in an unknown
  environment~\cite{cao2021tare}, where POMDP modeling is a natural
  choice. The PLGRIM~\cite{kim2021plgrim} framework follows the
  hierarchical approach to make planning over long time horizons
  feasible in this context. Locally, PLGRIM uses POMCP to cover the environment search
  space efficiently. An entropy based objective is used to
  reduce the uncertainty over possible
  environments. Globally, PLGRIM employs
  QMDP allowing for the robot to plan how to reach
  new areas. QMDP makes the strong assumption that
  after the first time step the environment is fully observable.
  This allows for efficient long horizon planning on the global level, but is appropriate only if belief
  uncertainty in the future can be ignored.
  Follow-up work~\cite{peltzer2022fig} avoids using QMDP at the global scale.}

\subsection{Autonomous driving}
\label{sec:autonomousdriving}

\begin{figure}
  \centering
  \includegraphics[width=\columnwidth]{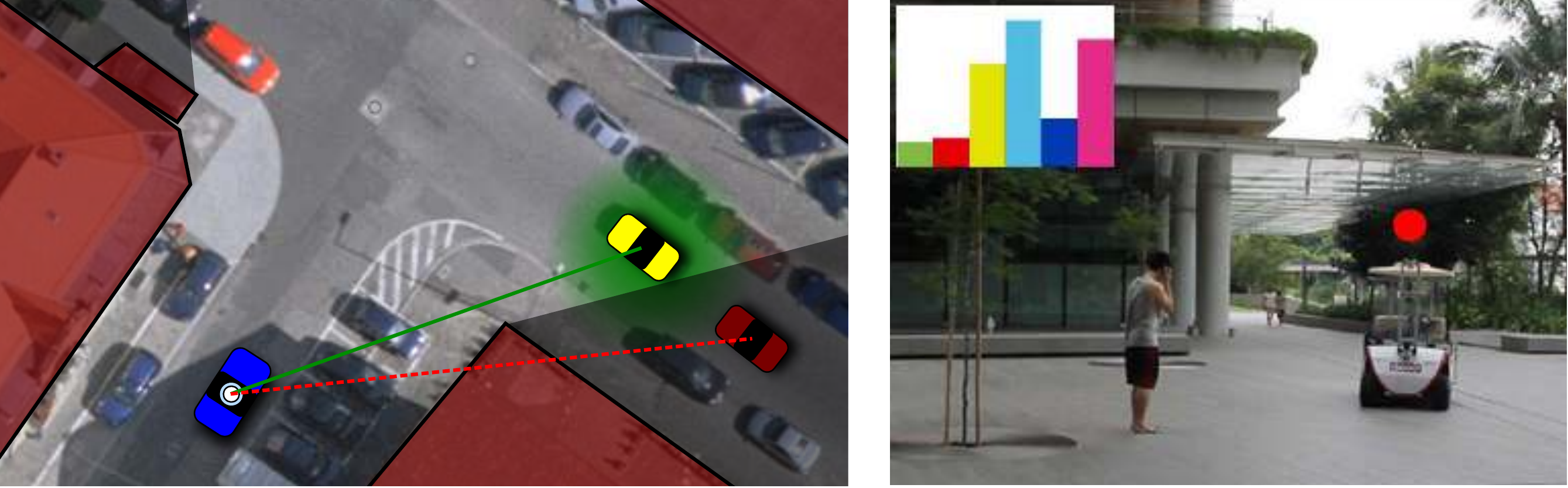}
  \caption{\textbf{Left:} In autonomous driving the agent has to typically reason
    about occluded vehicles and pedestrians (Figure
    from~\cite{brechtel2014probabilistic}). \textbf{Right:}
    {The histogram in the top left part shows the belief about intentions of human drivers} (Figure from~\cite{bai2015intention}).}
  \label{fig:autonomous_driving}
\end{figure}

One of the earliest success stories of autonomous driving was the
Autonomous Land Vehicle In a Neural Network (ALVINN) project at
Carnegie Mellon University~\cite{pomerleau1988alvinn}.
{ALVINN trained a neural network to map input images to discrete actions to follow a road.} Autonomous driving relies on state estimation,
predicting dynamic objects and controlling the car to prevent
accidents. In autonomous driving, a POMDP may cope with partial
observability due to occlusions caused by the environment and vehicles
and uncertainty of human driver and pedestrian intentions.
The ability in POMDP planning to reason how current actions affect future uncertainty is important, since a purely reactive control approach may not be able to anticipate a dangerous situation sufficiently early.

\paragraph{Key sources of uncertainty and challenges}
In autonomous driving, the key sources of uncertainty include other
traffic participants such as other vehicles or pedestrians and the
environment. The intentions of other traffic participants are commonly
partially observable. Parts of the static environment such as
buildings and dynamic objects such as other traffic participants
block parts of the view of the autonomous vehicle causing partial
observability. In autonomous driving, coping with the uncertainty is
crucial to prevent accidents.

Similarly to other tasks with navigation, hierarchical control is
common in autonomous driving. POMDPs are used in autonomous driving
for selecting high-level actions such as lane changing, distance
keeping, or overtaking other cars that facilitate safe and efficient
driving.  Low-level controllers for acceleration and steering are
responsible for the execution of these high-level actions. The
autonomous driving domain is closely related to the navigation domain
(see Section~\ref{sec:localization_navigation}) with continuous
states, action, and observations. Autonomous driving is also related
to target avoidance (see Section~\ref{sec:search_tracking_avoidance})
where other vehicles must be incorporated into the dynamics model to
predict and avoid collisions. Due to this the state space {must} be
augmented with information about other traffic participants, {often} including
their position and velocity \cite{wei2011point} or
their internal state including intention and aggressiveness
\cite{sunberg2017value}. Modeling of other traffic participants
connects autonomous driving with human-robot interaction {(see Section~\ref{sec:hri})}. The reward function
in autonomous driving often consists of multiple {possibly contradictory} objectives. The reward function often assigns negative
reward for collisions and positive reward for reaching a target
position.
Goals such as fuel efficiency can also be taken into
account.

\paragraph{Solution methods}
In the autonomous driving domain, {continuous state, action, and observation spaces are often discretized~\cite{thornton2018autonomous} to make computation tractable when applying discrete POMDP methods.}
Measures to allow for tractable solutions for complex
autonomous driving problems include short planning horizons, sparse
action alternatives, ruling out illegal or impossible actions,
mixed observability, and variable time granularity planning
\cite{ulbrich2015towards}.

Many POMDP approaches in autonomous driving focus on a specific subtask.
{A velocity controller for a car merging into traffic at a T-crossing is developed in~\cite{brechtel2014probabilistic}.}
The car
follows a pre-defined path, while its view of the environment may be blocked by objects such as buildings or other cars and trucks, see Fig.~\ref{fig:autonomous_driving} (left). 
The
position and orientation of dynamic vehicles are modeled as continuous
state variables. Observations specify the continuous state of perceived
dynamic objects and denote occluded objects as hidden. Velocity control is discretized.
{The belief state and value function are represented by using a set of samples, resulting in a dynamic discretization of the state and observation spaces.
}
Safe lane changing is addressed in~\cite{cunningham2015mpdm}, where the agent needs to take into
account the behavior of other cars. 
A state and observation
variable are {used} for each car resulting in large state and observation
spaces.
Instead of applying an approximate POMDP algorithm, two simplifying assumptions are used to tackle the resulting complexity: 1) a finite set of
known policies is sufficient for all vehicles, 2) vehicle dynamics and
observations are deterministic and evaluation using forward simulation
is sufficient.
These assumptions reduce the decision making problem into choosing from a set
of high-level behaviors.
{However, the scalability of this approach is limited as the number of possible behaviours grows exponentially in the number of traffic participants $n$.
With two possible policies per participant, there are $2^n$ possible combinations of policies to evaluate.}

{
The MODIA framework~\cite{wray2017,wray2021} proposes a solution for an autonomous vehicle to navigate an intersection.
Each external traffic participant is represented by a dynamically instantiated POMDP.
Actions for each POMDP are inputs to an executor module responsible for returning the action for the autonomous vehicle.
MODIA navigates intersections faster than a naive and cautious policy, with fewer unsafe situations than a baseline that does not model other traffic participants.
}

The intersection scenario where multiple human drivers arrive at an
intersection and the agent needs to decide when to proceed is considered by~\cite{song2016intention}. 
The agent predicts the intention of each human driver {to safely proceed across the intersection}. In
\cite{song2016intention}, the underlying model is a POMDP although a
POMDP {solver is not used}. 
Instead, a hidden Markov model {infers} the most likely
intention given continuous observations of the relative poses of each
human driver. 
{The agent selects the policy yielding the highest cumulative reward by simulating a predefined set of possible policies.}
This approach cannot take information gathering actions
into account{, and so does not perform well when there is a need to actively reduce uncertainty about the intention of the human driver prior to acting.} 
A collision avoidance scenario at an intersection is also considered in~\cite{hubmann2017decision}.
The agent controls car
acceleration using a POMDP while driving on a predefined path. The
state space contains the continuous states of all vehicles. The
observation space contains {a discrete path choice for each other car}.
{Aggregating small variations in trajectories to a set of path choices reduces the number of possible observations, allowing scaling to more complex intersections.}

Driven by the need to reduce the computational cost, a two-level hierarchical
planner is used in \cite{luo2018autonomous} and
\cite{bai2015intention} for intention-aware autonomous driving among pedestrians, see also Fig.~\ref{fig:autonomous_driving} (right). 
The
planner takes the global intention of pedestrians and their local
interactions into account. For efficiency reasons states with mixed
observability are used where the position and velocity
of the pedestrians are considered fully observable but their
intentions are only partially observable. To avoid collisions with
pedestrians, \cite{pusse2019hybrid} combines deep reinforcement
learning (DRL) with a POMDP solution. \cite{pusse2019hybrid} uses
DESPOT as the underlying POMDP algorithm which then uses state/action
value functions, that is, Q-functions from DRL as leaf node
approximations. {The authors in}~\cite{ychsu2020} model the decision making
of a car which is uncertain whether a pedestrian intends to cross a
street as a POMDP. The pedestrian being either ``reckless'' or ``cautious'' and
their hidden intent influence the transition probabilities. 
{The demonstrated strength of the POMDP approach is that the computed policy gathers information about the pedestrian's intent when it is considered valuable.
The policy could also leverage the autonomous vehicle's own actions to influence pedestrian behavior.
}

\subsection{Target search, tracking, and avoidance}
\label{sec:search_tracking_avoidance}

\begin{figure}[t]
  \centering
  \includegraphics[width=\columnwidth]{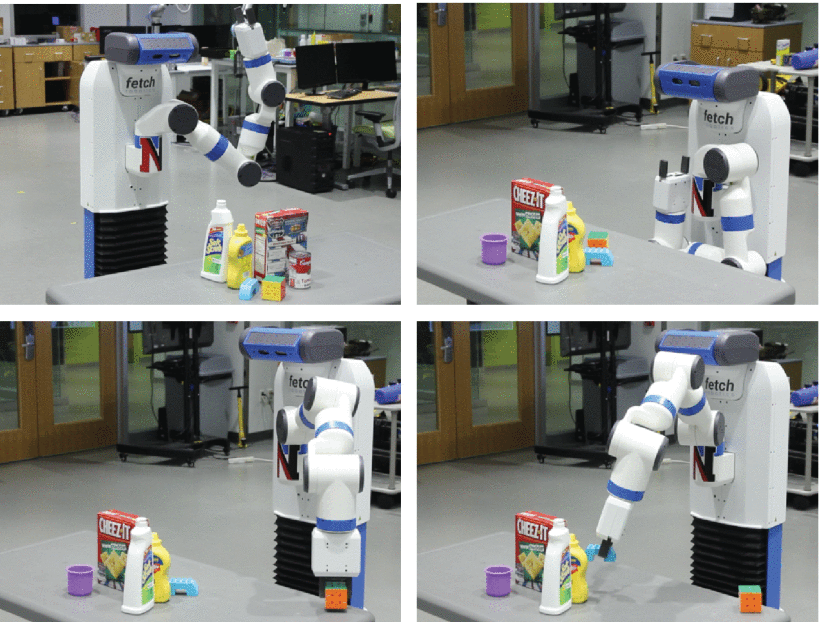}
  \caption{POMDPs are applied to plan sequential strategies for tasks
    such as object search {which require balancing between information
      gathering and exploiting current knowledge}.
      \textbf{Top left}: the robot is searching for the target light blue toy brick, but it is not visible. \textbf{Top right}: moving to a different vantage point, the robot finds the target object blocked by the multicolored cube. \textbf{Bottom left}: the robot has pushed the cube out of the way. \textbf{Bottom right}: The target object is grasped. (Figure from~\cite{xiao2019online})}
  \label{fig:search_tracking}
\end{figure}

Robotic applications involving target search, tracking and avoidance include
aerial photography, search and rescue, and
autonomous flight. While tracking or searching for a target, the robot also needs to navigate and localize, familiar tasks from
Section~\ref{sec:localization_navigation}. Furthermore, the robot needs to model target behavior and sensor uncertainty to reason about how to
gather more information about the targets (search, tracking; see Fig.~\ref{fig:search_tracking}) or how to
avoid the targets (avoidance). 

\paragraph{Key sources of uncertainty and challenges}
Target behavior is highly domain specific and difficult to model and predict, making it an important source of uncertainty in tracking, search, and avoidance problems.
Furthermore, targets may be tracked using sensors that produce high-dimensional observations, for example cameras~\cite{martinez2009bayesian,spaan2010active,sridharan2010planning,zhang2013active,atanasov2014nonmyopic,valimaki2016optimizing,wandzel2019multi,xiao2019online,wang2020pomp}.
It is challenging to determine probabilistic models for target behaviour and high-dimensional observations that are useful for planning while {being at the same time computationally feasible}.

The state of a single target is often discretized to a finite-dimensional grid, see, e.g.,~\cite{foka2003predictive,miller2009pomdp,capitan2013decentralized,satsangi2018exploiting}.
The number of grid elements equals the number of possible states the target may be in.
{If multiple targets are tracked}, the size of the state space is equal to the product of the individual grid sizes.
This leads to an exponential growth of the state space size in the number of targets, leading to computational challenges in solving the related POMDP.

\paragraph{Solution methods}
Hierarchical approaches including multiple levels of decision-making and control have been widely applied in target search, tracking and avoidance~\cite{foka2003predictive,sridharan2010planning,capitan2013decentralized,zhang2013active,vanegas2016uav,goldhoorn2018searching,yi2019indoor}, see also Section~\ref{sec:localization_navigation} for hierarchical approaches in localization and navigation.
POMDP models are applied to plan a strategy for tracking a target using an unmanned aerial vehicle in~\cite{vanegas2016uav}, while flight control is delegated to a low-level controller.
A POMDP planner produces waypoints for a robot in~\cite{capitan2013decentralized,goldhoorn2018searching}, deferring the traversal task to a lower level controller.
Hierarchical POMDPs for visual problem solving are suggested in~\cite{sridharan2010planning,zhang2013active}, and a tree with multiple levels of progressively finer-scale POMDPs is considered in~\cite{foka2003predictive}.
A hybrid approach that uses an MDP model when a target is visible and a POMDP model when not is proposed in~\cite{yi2019indoor}.
{Hiearchical methods can address long planning horizons by planning on a higher level of abstraction.}

Discretization of the target state allows the application of discrete POMDP solvers to tracking and search problems.
Grid-based discretization is used in, e.g.,~\cite{foka2003predictive,miller2009pomdp,capitan2013decentralized,satsangi2018exploiting}.
State space factorization is used for object oriented POMDPs in~{\cite{wandzel2019multi,zheng2021}}, and mixed observability -- the fact that some factors of the state may be fully observable -- is exploited in~\cite{ong2010planning,gupta2017decision}.

Others directly solve continuous-state tracking and search problems~\cite{brunskill2008continuous,martinez2009bayesian,bai2012unmanned,burks2019optimal}.
Multi-modality in dynamics that differ depending on the state is considered in~\cite{brunskill2008continuous}.
In~\cite{martinez2009bayesian} Bayesian optimization is applied to directly search for policies while simulation rollouts are used to evaluate policies.
For unmanned aerial vehicle collision avoidance with a continuous-state formulation, \cite{bai2012unmanned} proposes to use Monte Carlo value iteration.
Up to ten continuous state space dimensions in a search problem are handled by the recent variational method that parametrizes the POMDP using Gaussian mixtures~\cite{burks2019optimal}.

{Using} image data with POMDP methods has been proposed for object search~\cite{zhang2013active,wandzel2019multi,xiao2019online,wang2020pomp,zheng2021}, tracking~\cite{spaan2010active}, exploration~\cite{martinez2009bayesian,valimaki2016optimizing}, query answering~\cite{sridharan2010planning}, and object pose estimation~\cite{eidenberger2010active,atanasov2014nonmyopic}.
Instead of using images directly as observations in the POMDP which would be computationally prohibitively expensive, the images are abstracted into a low-dimensional representation.
For example, in~\cite{atanasov2014nonmyopic} a set of nominal object poses are applied as a basis for estimating arbitrary poses, in~\cite{xiao2019online} observations are formed by detecting objects by point cloud segmentation, and in~\cite{martinez2009bayesian} the landmark detections extracted from the image are used as observations.
{In~\cite{zheng2021}, a spatial abstraction is obtained via a hierarchical belief representation, where deeper levels represent the environment at a progressively finer resolution.
However, similar to~\cite{wandzel2019multi}, dependencies between objects are not considered, and a POMDP problem must be solved for each level of the hierarchy in parallel.
}

{Observations allow estimating the current belief which is then used
for decision making.} {A belief-dependent reward function} can be useful for search tasks where the objective is to reduce uncertainty.
For example, the entropy {of the belief state} quantifies the robot's uncertainty about the state.
{Negative entropy as reward has} been applied in the discrete state~\cite{hollinger2009efficient,zhang2013active,lauri2017multi,satsangi2018exploiting} and continuous state~\cite{eidenberger2010active,valimaki2016optimizing} settings.
Visual object search is addressed by the top level of the hierarchical method in~\cite{zhang2013active} using the policy gradient planner of~\cite{buffet2009factored}.
The greedy policy heuristic with simulation-based estimation of expected {reward} is used in~\cite{valimaki2016optimizing}.
Cooperative target tracking is treated in~\cite{lauri2017multi}, computing belief state entropy on a discretized representation projected from a continuous-state tracking filter.
Reward submodularity has been exploited in single-robot~\cite{satsangi2018exploiting} and multi-robot~\cite{hollinger2009efficient,corah2019distributed} tracking and exploration to develop approximation algorithms.
{The advantage of methods using submodularity is that they have performance guarantees for computationally cheap greedy policies.}

\subsection{Manipulation and grasping}
\label{sec:manipulationgrasping}

\begin{figure}[t]
  \centering
  \begin{tabular}{cc}
  \includegraphics[height=4cm]{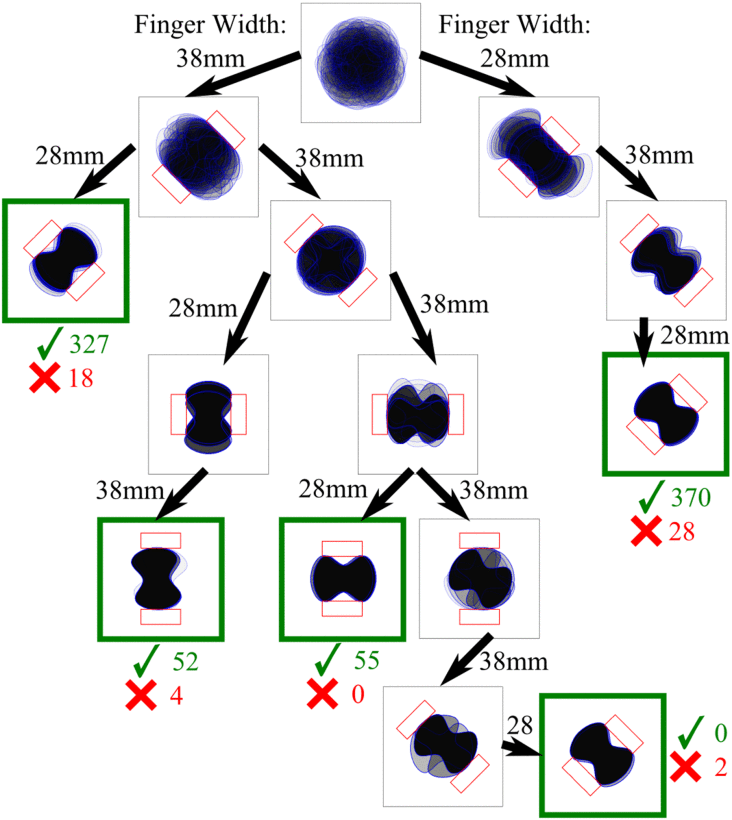}&
  \includegraphics[height=4cm]{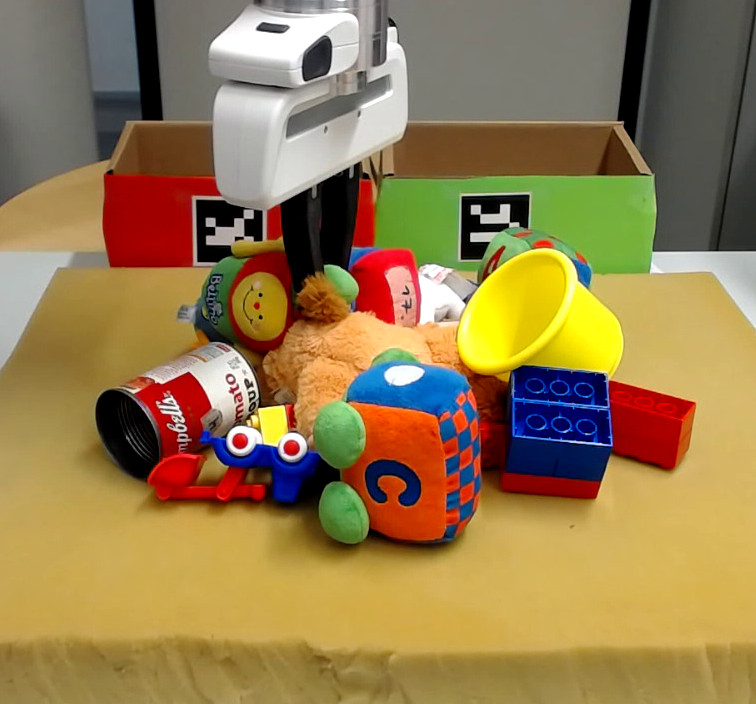}
  \end{tabular}
  \caption{\textbf{Left}: A policy for grasping an object in various configurations to reduce object shape and pose uncertainty (Figure from~\cite{zhou2017probabilistic}).
\textbf{Right:} In multi-object
     manipulation, occlusions and uncertainty on object identities may
     result in high partial observability~\cite{pajarinen2020pomdp}.}
  \label{fig:manipulation}
\end{figure}
Robotic arms have been used in the industry since the early 1960s
for manipulation and grasping in tasks such as assembly, polishing,
and painting. In fact, robotic manipulation is one of the cornerstones
of the modern manufacturing industry. In the classical industrial setting
uncertainty and partial observability is typically eliminated by exact
modeling of the task from start to finish. However, in unknown,
unstructured and cluttered environments, using noisy sensors, in collaboration with humans, or in
tasks with hard to model objects such as
textiles~\cite{monso2012pomdp} taking uncertainty and partial
observability into account is crucial and a POMDP is the de facto
model.

\paragraph{Key sources of uncertainty and challenges}
Robotic manipulation typically occurs in limited space that contains
multiple objects including one or more robot arms. {Cameras, static or mounted, and tactile sensors are often used for sensing.}
A key source of uncertainty are occlusions due to objects and robot arms partially blocking the view {of} objects. Another source of uncertainty are the a priori unknown physical properties and
identities of objects. The main challenge is how to manipulate
unknown objects with uncertain dynamics based on partially occluded
noisy sensor readings.

In more detail, a robot often manipulates objects on a flat surface
such as a table. The robot controls the joints of a robot arm or the
fingers of a robotic gripper.
POMDPs are used for task planning where high level actions indicate
which object to manipulate and what kind of manipulation to perform,
such as picking up or moving an object.
{Each object may have different attributes such as color and
  position that define the state of a single object. However, the
  number of different combinations of object attributes, and thus the
  whole state space, grows exponentially with the number of objects.}

Observations are typically recorded by imaging sensors
observing the robot's workspace. Due to the high dimensionality of
image input, the images are usually preprocessed and mapped to
individual objects. In manipulation, partial observability is a result
of noisy visual or tactile sensors and occlusions in the environment.

\paragraph{Solution methods}
In robotic \textbf{grasping}, especially when based on tactile sensors
\cite{hsiao2007grasping,koval2016pre,zhou2017probabilistic}, the poses of the grasped
object and the gripper are partially observed. To handle the naturally
continuous action and state spaces, in early work using
POMDPs for robotic grasping with tactile feedback, e.g.,~\cite{hsiao2007grasping}, a model of the environment is used where movements of the robot hand are constrained to a discrete set and
the state space is divided into discrete cells. {The agent is able to compute a policy using a binary
observation signal from its tactile sensor}.
{Scaling to larger problems was later demonstrated by selecting trajectories from a candidate set~\cite{hsiao2008robust}.}
In~\cite{zhou2017probabilistic}, an offline planner solves for a tree policy used online to infer the shape of an object to be grasped based on tactile data, see Fig.~\ref{fig:manipulation} (left).

Aligning the fingers of a
multi-fingered robot hand and non-prehensile manipulation is {used} in~\cite{horowitz2013interactive}. 
The robot pushes a poorly observable object such as a small nut on a
table to localize it, after which it can be grasped. 
To compute a policy, SARSOP is applied on a specifically designed state space where the
continuous 2-D location of the object is
discretized finely around the estimated object location. Further away from the estimated object location
the location is only coarsely discretized to represent the relative displacement from the current object location estimate.

Taking the full grasping motion into account allows for moving the
robot hand close to an object and then locating and grasping the
object using tactile sensors. 
In~\cite{koval2016pre}, the grasping motion is split
into a pre-grasp open-loop movement which is
optimized using the A*-algorithm, and into a post-grasp closed-loop
POMDP based push-grasping approach which {uses} partially observed
tactile sensors. It is shown that the full POMDP solution
outperforms an approach based on the QMDP heuristic, suggesting a closed loop policy is needed to be able to gather necessary
information during the task.

A POMDP model for tasks where parts of {the} system dynamics are fully observable but where state
variables are either constant or change deterministically is presented in~\cite{chen2016pomdp}. 
The model corresponds to choosing a fully-observable MDP according to a hidden discrete
state variable. The model is evaluated in a simulated robot manipulation task where the robot needs to grasp a cup
with an uncertain initial position on a planar surface using binary
tactile sensors. The position of the robot arm is fully observable and
deterministic. The approach discretizes actions, observations, and
states.

In another line of grasping research,
POMDP solvers and deep learning are combined~\cite{garg2019learning}. First, data are generated by planning grasps on a set of objects using the DESPOT
POMDP algorithm. Second, an imitation policy that mimics the behavior of the POMDP
policy is learned in a supervised manner.
It is empirically shown that the imitation policy generalizes to novel objects not present in the training data.

\textbf{Multi-object manipulation.} Multi-object manipulation {requires reasoning about multiple objects and their interactions concurrently}~\cite{monso2012pomdp,pajarinen2017robotic,pajarinen2014robotic,li2016act,xiao2019online,pajarinen2020pomdp}. In
multi-object manipulation, a POMDP model takes into account partial
observability due to sensor noise and objects occluding other
objects. Similarly to target search, tracking, and avoidance in
Section~\ref{sec:search_tracking_avoidance} a main challenge in multi-object
manipulation is the exponential growth of the state space with the
number of objects. {Techniques such as image rendering or physics simulations are available to model observations and scene dynamics, but it is challenging to find models that are amenable to a POMDP formulation while remaining computationally tractable in practice.}

One approach is to approximate states, actions, and observation with
discrete values by taking advantage of the inherent properties of the
actual task such as cloth separation \cite{monso2012pomdp}. A cloth may be removed when it is not entangled with other clothing items. A state space model that counts the
number of clothes in two different areas on a table is used, resulting in a
POMDP with few states, observations and actions. There are two actions for removing clothes from the table and $20$
actions for moving objects from one pile to another chosen using
different robot hand finger configurations and visual features.

Other works such as~\cite{pajarinen2017robotic,pajarinen2014robotic,li2016act,xiao2019online,pajarinen2020pomdp}
model each object as a separate state
variable. A robot
observing several objects resting on a table with an RGB-D sensor is considered in~\cite{pajarinen2017robotic}. Each state variable specifies the
properties of an object such as color or location. An action lifts or
moves an object. Observations include information about properties of
the objects behind the moved or lifted object. The
RGB-D scene is segmented into objects, and the amount of mutual occlusion is estimated.
The occlusion information is applied to estimate grasp success
probabilities and observation probabilities for the POMDP model to be solved.

In \cite{li2016act}, the goal is to find objects from a set of
possible known objects resting on a shelf. The
location on the planar surface and the orientation of the objects is
discretized. Discretized actions allow moving an object {onto} the
shelf, {off} of the shelf, and tagging the target object. Observations
provide the properties of each object in the scene. Similarly to
\cite{pajarinen2017robotic}, observation probabilities depend on the
amount of occlusion which is estimated by
projecting objects on top of each other.

Object search using a POMDP model that also considers completely occluded objects is tackled in~\cite{xiao2019online}.
The correct segmentation of known object models is assumed to be known, with some uncertainty in
object locations. Actions and observations are discretized. Due to the
large state space a particle representation is used for the belief, similar to
\cite{pajarinen2017robotic,li2016act}. To
generate each belief particle \cite{xiao2019online} selects object
locations from the segmented scene and adds sampled noise. Fully
occluded objects are sampled in the occluded volume behind visible objects. A variant of
POMCP is used to compute a policy.
Fig.~\ref{fig:search_tracking} shows an example execution of the search policy.

With cluttered unknown objects the RGB-D segmentation of a scene and the
number of objects is uncertain, see Fig.~\ref{fig:manipulation} (right). Scene uncertainty is incorporated into the POMDP belief in~\cite{pajarinen2014robotic,pajarinen2020pomdp}, allowing planning of manipulation actions over object hypotheses. Each particle in the belief representation
corresponds to a specific set of hypothetical objects. The robot may attempt to
grasp and move a hypothetical object. The outcome of the grasp attempt provides information about the actual objects to update the belief. 

\subsection{Human-robot interaction}
\label{sec:hri}

\begin{figure}[t]
  \centering
  \includegraphics[height=3.5cm]{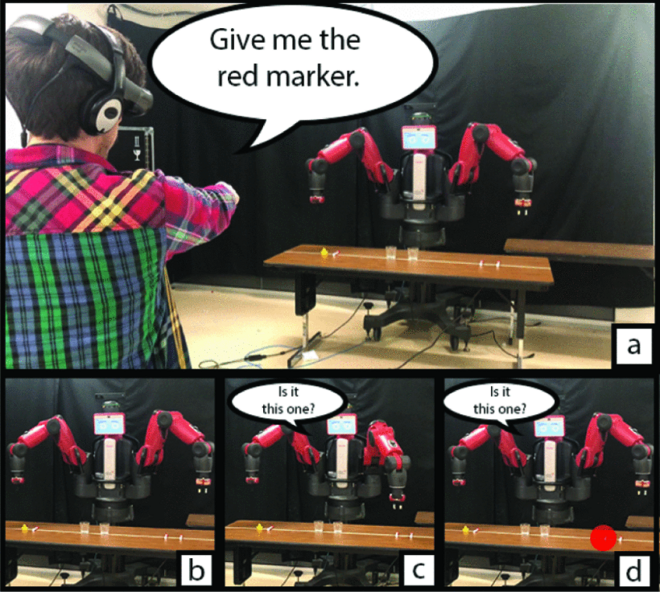}~\includegraphics[height=3.5cm]{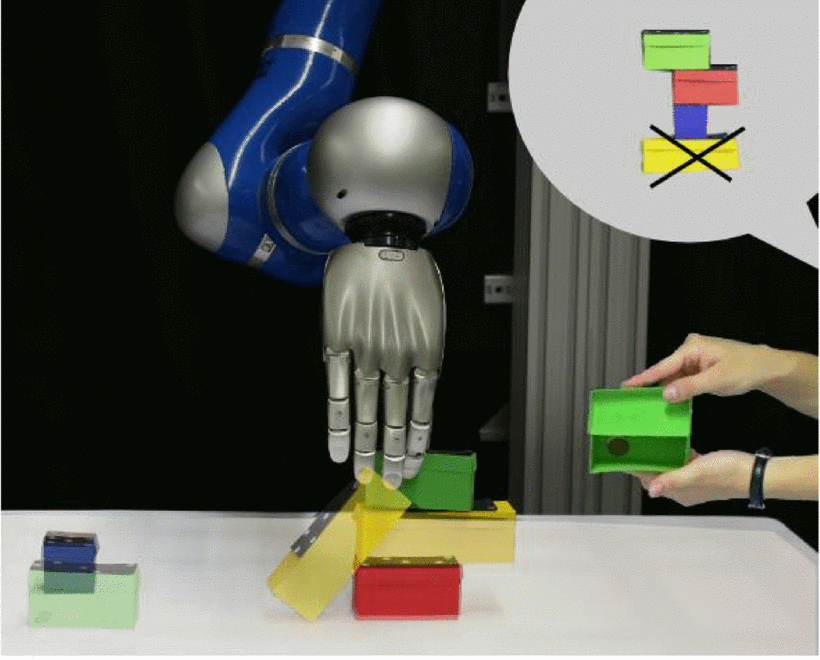}
  \caption{\textbf{Left:} Human-robot interaction uses POMDPs to infer human intent from multimodal interaction data such as gaze, speech, and gestures (Figure from~\cite{rosen2020mixed}). \textbf{Right:} Planning bi-directional communication with a human (Figure from~\cite{hoelscher2018utilizing}).}
  \label{fig:hri}
\end{figure}

Human-robot interaction (HRI) is {necessary in domains} ranging from industry to healthcare. In
HRI, a robot typically needs to reason about possible human behavior
in order to perform the assigned task efficiently. 
No sensor exists to directly measure human intentions or mental state, which must therefore be inferred from other data.
Furthermore, humans may make decisions based on different
sensory input than the robot. POMDPs are used in HRI to make decisions
with these kinds of uncertainties in a principled way, see Fig.~\ref{fig:hri} for examples. The ability to model information gathering allows POMDPs to be used in
tasks requiring communication and other kinds of interaction which can
be crucial in HRI.

\paragraph{Key sources of uncertainty and challenges}
A key source of uncertainty specific to HRI is the partially
observable human state or intention.
{A model of human behaviour can help} ensure safety and human comfort while {the robot performs its} tasks. Such models often consider quantities related to the mental state
of a human, characteristics such as competence level, and the physical
pose and location of a human. The mental state of a human is often
defined in terms of discrete states and connected to human models in
psychology~\cite{hoey2010automated,taha2011pomdp,chen2018planning}. In
physical human-robot interaction, the pose and location of a human is
inherently continuous valued~\cite{wang2017anticipatory} and may
require further approximation techniques. While presence of humans
increases modeling complexity, humans can also provide information to
the robot through communication. By considering communication as a
means to gather information, POMDPs are applied to reason about
and plan
communication~\cite{armstrong2007oracular,atrash2009bayesian,atrash2010bayesian,hoelscher2018utilizing,unhelkar2020decision}.

\paragraph{Solution methods}
In early work on using POMDPs for decision making in HRI, {the robot assistant Pearl was deployed in a nursing home~\cite{pineau2003towards}.
Pearl navigated the nursing home, guiding residents to appointments while answering queries, e.g., concerning the current weather.
}
Due to the intractable
complexity of exact POMDP solutions~\cite{cassandra1997incremental},
Pearl {factors its belief into probability distributions over the}
location and status state variables. For computational efficiency,
the hierarchical POMDP chooses a combination of high and low level actions
at each time step.
Pre-defined high-level discrete actions such as ``Inform'' or ``Move'' are each associated with a set of lower level actions such
as ``SayTime'', ``SayWeather'', or ``VerifyRequest''.
{Pearl demonstrated the feasibility of POMDPs for high-level decision making in an autonomous robot assistant. Taking into account the needs and capabilities of human partners, e.g., adapting to different walking speeds, was noted to be crucial for successful interaction.}

Several POMDP models incorporate latent state variables representing
the mental state of the
user~\cite{hoey2010automated,taha2011pomdp,chen2018planning}.
{Latent state variables can capture both long and short term
  mental states. In \cite{hoey2010automated}, POMDPs are used for
  monitoring the mental state of elderly people with disabilities and
  providing assistance in tasks such as hand washing. The system
  tracks the hand washing of people with dementia and displays
  messages to the user when needed. The POMDP model consists of
  discretized observations, a discrete action set of possible
  messages, a special action for calling a caregiver, and latent state
  variables that reflect both mental state and hand washing
  phase. Mental states can be also used in
  shared-control. The POMDP HRI framework in~\cite{taha2011pomdp} uses
  the independence between certain mental states for an efficient
  algebraic decision diagram~\cite{poupart2005exploiting} based
  POMDP implementation enabling shared-control in a wheelchair
  task. While modeling the internal state of a
  human~\cite{hoey2010automated,taha2011pomdp} can yield a
  comprehensive model, a simpler approach is taken in~\cite{chen2018planning} to model human
  trust level in a single latent state variable. The dynamics for the
  state variable are learned from user interaction data which allows
  predicting when the human would intervene with the robot's table
  clearing task.}

{In contrast to mental state models, in~\cite{wang2017anticipatory} the robot models the user as part of a table tennis environment and predicts the current target of the ball for the user's hitting motion.} 
The dynamics
and observations are modeled using a Gaussian Process. The robot
executes a binary action at each time step to either initiate a table
tennis movement or to wait. In \cite{wang2017anticipatory}, the
methods {used} for POMDP computation include Least-Squares
Policy Iteration (LSPI) and a model-based random shooting Monte-Carlo
method.

Human-robot communication can be crucial in HRI.
Some works plan when to ask humans for further
information~\cite{armstrong2007oracular,rosenthal2011modeling,hoelscher2018utilizing}
while others focus on bi-directional
communication~\cite{rosen2020mixed,unhelkar2020decision}. 
In the Oracular POMDP (OPOMDP) model~\cite{armstrong2007oracular} the robot may ask a human for full
state information at a cost, but does not otherwise receive
information. 
Computational speed-ups are achieved by splitting the task into information gathering oracular actions and
domain-level actions.
The OPOMDP model is extended by~\cite{rosenthal2011modeling} to
Human Observation Provider POMDPs (HOP-POMDPs) where human answers are
stochastic. 
The POMDP of~\cite{rosen2020mixed} models human-robot interaction via speech, gestures, and eye gaze.
Instead of asking for full state information, in
\cite{hoelscher2018utilizing} the robot asks task-specific questions, such as information about an object. 
The task goal is specified as a set of logic sentences by a human using a
graphical user interface. 
In~\cite{hoelscher2018utilizing}, the
PPGI planner~\cite{pajarinen2017robotic} is extended to
maximize the probability of satisfying the logic sentences.
In a shared workspace task with a latent state model for the human teammate's behavior, \cite{unhelkar2020decision} uses a POMDP to decide which type of communication the robot should engage in.
The approach relies
on a predefined communication cost model, a human response model, and
a human action-selection model with mental
states. For policy optimization, \cite{unhelkar2019semi,unhelkar2020decision}
modify the Regularized-DESPOT
algorithm~\cite{somani2013despot}: planning is performed during action execution
for all possible observations, and the actual perceived
observation is then used to choose the next action.

In HRI and in POMDP based robotics in general, the POMDP model is
typically specified by an expert either fully or partly. In
\cite{atrash2009bayesian,atrash2010bayesian}, the robot learns parts
of the POMDP model from a user. 
Cases with an unknown reward function~\cite{atrash2009bayesian} or unknown observation model~\cite{atrash2010bayesian} have been considered.
Both works~\cite{atrash2009bayesian,atrash2010bayesian}
represent the distribution over POMDP parameters drawn from a prior
distribution as a finite set of hypothetical POMDPs. At each time step
the user provides an action which is executed. The robot updates its
set of POMDP model hypotheses to better match the provided discrete-valued action~\cite{atrash2009bayesian} or observation~ \cite{atrash2010bayesian}.
The approaches are evaluated on a robotic wheelchair
control system. Instead of assuming predefined state and observation
spaces, \cite{zheng2018pomdp} learns from human-robot collaboration
data the POMDP state and observation spaces using non-parametric
probability distributions, and, maximum likelihood transition and
observation probabilities. 
Model confidence intervals are used to estimate a lower bound for the amount of data required to
reach a specific POMDP control performance level.

Other tasks where POMDP based HRI has been studied include human-robot
social interaction~\cite{schmidt2010learning,broz2011designing},
medical diagnosis~\cite{petric2019hierarchical}, and
assistive tutoring of students~\cite{ramachandran2019personalized}.
In human-robot social
interaction, \cite{broz2011designing} uses a POMDP to optimize robot
social interaction in the task of yielding or going first in a driving
simulator. The state space consists of discrete information about
human intention, the two cars and the environment. The dynamics model
is learned from simulations. In other work, \cite{schmidt2010learning}
investigates human-robot social interaction by observing
human-human-interaction. Contrary to typical POMDP based HRI, in
\cite{petric2019hierarchical} the POMDP decides on the next task in
autism spectrum disorder (ASD) diagnosis instead of on high or low
level actions. In \cite{petric2019hierarchical}, each task is an
interaction sequence between a robot and a child.
\cite{ramachandran2019personalized} defines a discrete valued POMDP to tutor
4th grade students in mathematics. The robot chooses 
a tutoring action before the student attempts a math problem
and observes the speed and accuracy of the student. Positive {reward is} accrued
for increased student knowledge, engagement level, and for fast progress.

\subsection{Multi-robot coordination}
\label{sec:multirobot}

\begin{figure}[t]
  \centering
  \includegraphics[width=0.7\columnwidth]{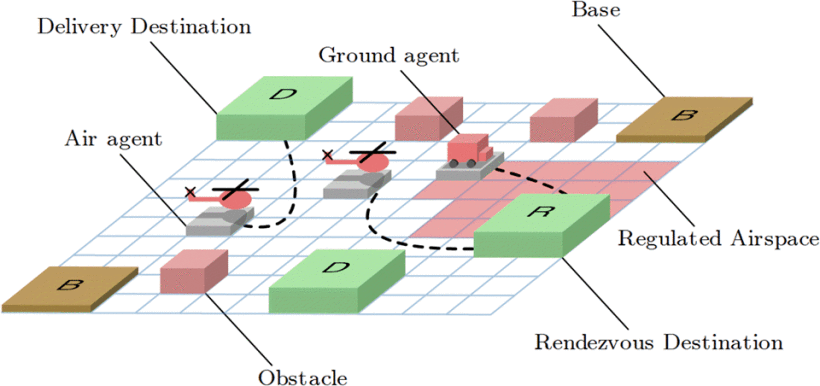}\\
  \includegraphics[width=0.7\columnwidth]{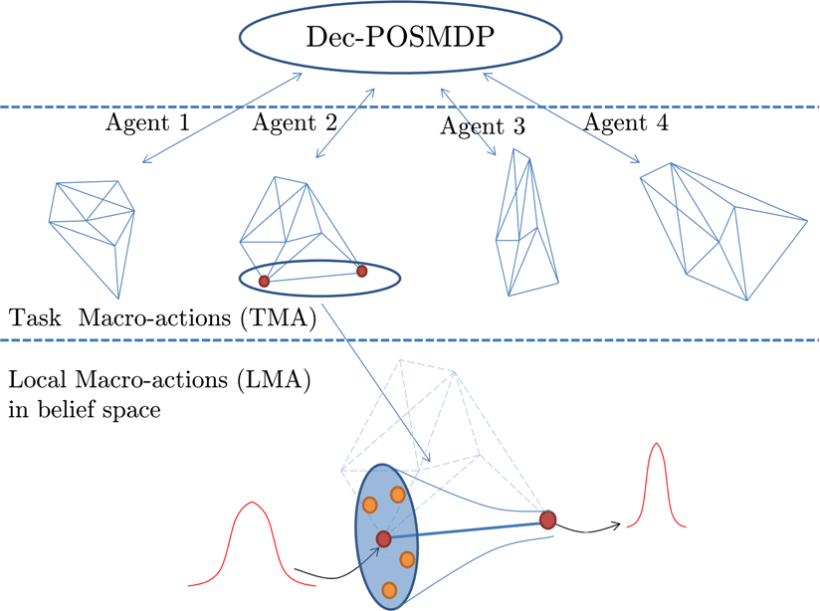}
  \caption{\textbf{Top:} Multi-robot logistics domain with complex interactions under uncertainty. Air and ground agents cooperate to deliver packages to their destinations. \textbf{Bottom:} The hierarchical multi-robot planning method of~\cite{omidshafiei2015decentralized,omidshafiei2017decentralized} consists of top-level task macro actions (TMAs) that correspond to specific tasks (e.g., picking up a package). Each TMA is a set of local feedback controllers (LMAs) that are executed to transform the initial beliefs of the TMA (shaded region at the bottom) to a goal belief (red circle). Figures from~\cite{omidshafiei2015decentralized}.}
  \label{fig:multi_robot}
\end{figure}

Multi-robot coordination is useful in applications such as target tracking or logistics (see, e.g., Fig.~\ref{fig:multi_robot} (top)).
In the following, we outline challenges specific to the multi-robot setting typically not present in single-robot cases, and review recent advances in POMDPs for the multi-robot setting.
We consider \emph{cooperative} settings with a shared goal, and do not address competitive or adversarial settings.
{We do not consider multi-robot task allocation, see, e.g.,~\cite{korsah2013comprehensive} for a survey.}

\paragraph{Key sources of uncertainty and challenges}
The key source of uncertainty in multi-robot coordination is the asymmetry of information available to each robot in the team.
If communication is limited, the action-observation history of robot $j$ is in general not known to another robot $i$.
This means that it is not possible for $i$ to predict which action $j$ will take, as this depends on both $j$'s policy and $j$'s private action-observation history.
This makes coordination challenging.

{One way to avoid the complexity of a decentralized solution is to assume that robots can freely communicate so-as to share a single shared belief state~\cite[Sec.~2.4.3.]{oliehoek2016concise}. In the resulting multi-agent POMDP} the action space consists of joint action tuples $(a^1, a^2, \ldots, a^n)$ of each of the $n$ robots in the team, and the observation space consists of joint observation tuples $(z^1, \ldots, z^n)$.
{This multi-agent POMDP is simply a POMDP where the actions and observations have a tuple structure, and standard POMDP solvers may be applied to compute a policy.}
{Such a policy is often realized by deploying} a centralized coordinator that uses the current belief state to determine the next individual action $a^i$ of each robot $i$, and then receives the individual observations $z^i$ to update the belief state via Bayes' rule.
This approach has been successfully applied in many multi-robot coordination problems, see, e.g.,~\cite{spaan2008cooperative,spaan2010active,ahmadi2019safe,corah2019distributed,viseras2019deepig,yi2019indoor,bhattacharya2020multiagent}.
The communication requirement is a major practical challenge of applying multi-agent POMDPs.

A \emph{decentralized POMDP} (Dec-POMDP) takes the approach of ``centralized planning with decentralized execution''~\cite{oliehoek2016concise}.
Like the multi-agent POMDP, the Dec-POMDP also replaces the action and observation spaces by the joint action and joint observation space.
However, the solution of a Dec-POMDP is a \emph{decentralized} policy that each robot in the team can execute individually, without having to know the action-observation histories of other robots{, or sharing a belief state}.
The policy is computed centrally, and then distributed to each robot for decentralized execution.
No communication at execution time is assumed, but any available communication can be explicitly modeled~\cite[Sec.~8.3]{oliehoek2016concise}.
This approach is used for multi-robot coordination in, e.g., ~\cite{omidshafiei2015decentralized,omidshafiei2017decentralized,amato2016policy,lauri2017multi}.
While the Dec-POMDP approach is extremely {general}, it is also computationally more challenging than solving a POMDP~\cite{bernstein2002complexity}.
We refer the reader to~\cite{oliehoek2016concise} for technical details.

\paragraph{Solution methods}
As mentioned above, if the robots in the team can always communicate with each other, coordination can be reached by solving a multi-agent POMDP, see, e.g.,~\cite{spaan2008cooperative,spaan2010active,ahmadi2019safe,corah2019distributed,viseras2019deepig,yi2019indoor,bhattacharya2020multiagent}.
The solution methods do not differ from the single-robot case.
Therefore, we focus here on two distinct approaches to planning in multi-robot coordination problems with limited communication.
First, a Dec-POMDP can be solved offline to agree on a joint strategy before plan execution.
Then, no communication at the time of policy execution is needed.
The Dec-POMDP is either solved directly, or it is decomposed into simpler single-agent planning problems solved locally.
Secondly, methods that use local communication during plan execution can reach a consensus, e.g., on a shared belief state, and use this to coordinate actions online.

A Dec-POMDP modeling a multi-robot task can be solved approximately~\cite{ragi2014decentralized} or exactly~\cite{lauri2017multi}.
In~\cite{lauri2017multi}, a multi-robot target tracking problem with a reward dependent on the robot team's joint belief is considered.
The theoretical foundations of such multi-robot information gathering were investigated in~\cite{lauri2020multi,lauri2020multib}.
Due to the computational complexity of Dec-POMDPs, these approaches are limited to small problems with a handful of agents, actions, and observations.
Alternatively, the \emph{semi}-Markov decision process setting allows temporal abstraction by modeling varying durations of actions~\cite{amato2016policy,omidshafiei2015decentralized,omidshafiei2017decentralized}.
This abstraction improves scalability by introducing additional levels of hierarchy through macro-actions composed of simpler closed-loop controllers, see Fig.~\ref{fig:multi_robot} (bottom).
Each high-level macro-action for a task executes lower-level closed loop controllers that transform an initial belief to a goal belief where the task is completed.
In~\cite{omidshafiei2017decentralized}, Gaussian beliefs are used at the lowest level, allowing application of macro-actions in continuous problems.
The problem is solved by Monte Carlo and cross-entropy based algorithms.
The method is demonstrated in a package delivery domain with four quadrotor robots with no online communication.
The completion times of macro-actions such as ``go to delivery location $j$'' are probabilistically modeled in the semi-Markov framework.
{The MacDec-POMDP framework, see, e.g.,~\cite{amato2016policy}, further removes the restriction that the completion time step of a macro action is determined before execution.}

Some works map the Dec-POMDP problem to a set of local single-agent problems that are easier to solve~\cite{capitan2013decentralized,zhang2017role,matignon2012coordinated,long2018towards}{, at the cost of losing optimality}.
In~\cite{capitan2013decentralized}, each agent plays one of {several} possible roles corresponding to a specific
local reward function.
At run-time, each agent solves a multi-agent POMDP to compute the value of each potential role. 
An auction algorithm {allocates} tasks to best-performing agents.
The auction requires communication between robots, but overall the method avoids the computational burden of solving a Dec-POMDP. 
The method of~\cite{capitan2013decentralized} is demonstrated in robotic environmental monitoring and cooperative tracking.
Role-based POMDP abstractions are also proposed in~\cite{zhang2017role}, where agents leave clues that are observable to the other agents, allowing inference of the roles and future actions of others without direct communication.
In~\cite{matignon2012coordinated}, a homogeneous team of robots explores an environment. The discrete Dec-POMDP problem is approximated as a set of local MDPs.
If communication is not possible, a robot acts independently by solving its local MDP.
Robots interact via communication with nearby other robots, affecting each others' local MDP value functions.
In~\cite{long2018towards}, multi-robot navigation and collision avoidance is cast as a set of POMDPs, one for each homogeneous robot in the team. A single shared policy is learned, which still induces individual behaviour as each agent conditions its actions on its individual observations.

When robots communicate during policy execution they can reach a consensus, e.g., by communicating and updating local beliefs~\cite{zhang2013active,capitan2013decentralized} or observations~\cite{ragi2014decentralized,goldhoorn2018searching}.
In case of communication breaks, missing data from other team members can be predicted based on the model parameters, see, e.g.,~\cite{matignon2012coordinated,capitan2013decentralized}.
Alternatively, coordination is achieved in~\cite{best2019dec} by periodically communicating policies with other robots.
Notably, no centralized planning phase is assumed, {which makes} the method fully distributed. 
Each robot locally plans a best response policy using MCTS, simulating other robots' actions according to the received policies.
The method of~\cite{best2019dec} was applied to agricultural multi-robot fruit detection in~\cite{sukkar2019multi}.
{A limitation of this approach is that it searches for open loop policies, potentially with a lower performance compared to an optimal solution.}
The approaches above allow a robot to continue executing its policy even if communication is interrupted.

\section{Challenges and Future Directions}
\label{sec:open_questions}
POMDPs have succeeded in many robot tasks that require reasoning about uncertain future outcomes using incomplete or noisy data.
This section outlines some of the outstanding challenges that we believe crucial for further progress in the field.
Accurate models are required for planning, but difficult to acquire.
Robotics tasks are often inherently continuous, while the classic POMDP framework has mainly focused on the discrete case.
Finally, in safety-critical {robotic tasks}, more robust probabilistic guarantees are needed to guarantee safety.

\textbf{Model uncertainty.}
In many realistic {robotic tasks} that could be modeled as a POMDP, there is considerable uncertainty about the model parameters.
Most approaches reviewed in this survey assume that
dynamics, observation, and reward models are given. However,
obtaining such models either by learning or  expert design can be
challenging in robotics. Kinematic modeling of rigid robots is well
understood but efficient modeling of heavy hydraulic
machines~\cite{xu2016review}, soft robots~\cite{kim2021review}, or,
complex physics in man-made and natural
environments~\cite{boonvisut2012estimation} are unsolved active
research topics.
{POMDP approaches often require model simplifications for tractability, and handling problems with intricacies such as
non-stationary or switching-mode dynamics may also be challenging.}

{Recent works have proposed several learning methods for robotic tasks that can help tackle model uncertainty, for example combination of planning methods with imitation learning for grasping~\cite{garg2019learning}, representation learning for beliefs~\cite{Nguyen20}, and multi-robot reinforcement learning for navigation~\cite{Wang20model} or temporal abstraction~\cite{Xiao20}.}
{Techniques such as Bayesian reinforcement learning (BRL)~\cite{ghavamzadeh2015bayesian} could be applied when} the model has been identified up to a handful of unknown parameters.
BRL treats the unknown parameters as additional state variables.
After defining a prior on the unknown parameters, a policy that optimally gathers information on the parameters is solved for~\cite{BaiHsu13a,ross2008bayesian,slade2017simultaneous}.
When considerably less prior information about the model parameters is available, e.g., when identifying a model for image observations, models could be  learned directly from data and then used for planning.
A promising recent trend is task-driven end-to-end learning, which integrates planning with model learning.
For example, we could learn latent space models for planning with image data~\cite{hafner2019learning} and for long-horizon tasks~\cite{schrittwieser2020mastering}. We could also {train a neural network to produce plans}, thus making the planning algorithm differentiable. The Differentiable Algorithm Network (DAN) leverages the latest advances in deep learning to learn a model most useful for planning, even though it may deviate from the ground truth~\cite{KarMa19}.
Translating these advances to robotic tasks is an important future direction.

\textbf{Continuous state, action, observation, and time.} 
The classic POMDP model consists of discrete states, actions, observations, and time, while robots operate in the physical world and often require continuous models for control. 
Discretizing continuous values can
result in inaccurate solutions especially in high dimensional
tasks. One approach is to assume specific probability distributions,
policies, or, state, action, observation representations. For example,
linear quadratic Gaussian control assumes Gaussian distributions and a
linear feedback control policy~\cite{van2012efficient}. 
Modern online POMDP algorithms, specifically, POMCP~\cite{silver2010monte} and DESPOT~\cite{somani2013despot}, relax these restrictive assumptions. They directly handle continuous states and observations through Monte Carlo sampling. However, the actions remain discrete. 
One future direction
is to transfer recent advances in
continuous POMDP methods~\cite{sunberg2018online} to novel robotic
tasks. 
Practical real world control often assumes
discrete time steps, even under full observability. However, some
systems can benefit from continuous time control, in robotics especially
non-linear systems that require fast responses. Continuous time MDPs have been
investigated as semi-MDPs~\cite{sutton1999between} which
consist of a sequence of continuous time fragments.
While existing work on continuous time POMDPs is scarce \cite{chaudhari2013sampling},
novel continuous time contributions
could enable solutions to new {robotic tasks}.

\textbf{Safe POMDPs in robotics.} Safety is an inherent requirement in
several robotic applications {such as human-robot interaction or autonomous
driving} where partial observability plays an important role. However, typically safety requires constrained POMDP
solutions~\cite{thiebaux2016rao,lee2018monte} that can guarantee how
often unsafe events may happen on
average~\cite{charnes1963deterministic,undurti2010online,thiebaux2016rao,lee2018monte}. 
{There is an opportunity for scaling and extending constrained POMDP methods
such that they become a standard tool in robotics.}
At
the moment these methods are however limited to {only a} few robotic
applications.
{On the one hand, these limitations are due to computational complexity, such as the ``curse of dimensionality'' due to state space discretization~\cite{wang2018bounded}, that limit the scale of problems that can be addressed.
On the other hand, safety-constrained methods are safe with respect to the POMDP \emph{model} of the task, whereas there may still be a gap between the model and the real task.}

{Robots are systems where knowing which sensors and actuators are fully operational is crucial for safety and robustness.
Such questions are tackled by system health monitoring, which is becoming increasingly unified with decision-making, see~\cite{balaban2019unifying} and references therein.
This implies further possibilities for leveraging POMDP planning for objectives such as greater fault resiliency in robotics.
}

\section{Conclusion}
\label{sec:conclusion}
Uncertainties in action effects, sensor data, {and environment states} are inherent to robot systems operating in the physical world and pose significant challenges to their robust  performance. 
In this survey, we have reviewed the POMDP framework for  robot decision-making under uncertainty, including both efficient algorithms for POMDP planning and  their applications in robotics. 
The success of POMDP planning on a wide range of different robot tasks clearly demonstrate its effectiveness and generality for robot decision making under uncertainty. Currently, 
sampling-based online POMDP planning algorithms stand out as a favored choice because of their scalability in high dimensional spaces and  ability to handle dynamic environments naturally.
Dealing with continuous state and action spaces, planning with high-dimensional sensory data, and acquiring accurate dynamics and sensor models are among the key challenges shared among many application areas of POMDPs in robotics.

In future work, we  expect to see tighter integration of POMDP model learning and planning. 
Further, by establishing a strong link to the theoretical guarantees provided by state-of-the-art POMDP solution algorithms, these new results will contribute to reliable and safe  decision making under uncertainty in many {robotic} systems. 

\ifCLASSOPTIONcaptionsoff
  \newpage
\fi

\bibliographystyle{IEEEtranS}
\bibliography{root}

\begin{thebibliography}{100}
\providecommand{\url}[1]{#1}
\csname url@samestyle\endcsname
\providecommand{\newblock}{\relax}
\providecommand{\bibinfo}[2]{#2}
\providecommand{\BIBentrySTDinterwordspacing}{\spaceskip=0pt\relax}
\providecommand{\BIBentryALTinterwordstretchfactor}{4}
\providecommand{\BIBentryALTinterwordspacing}{\spaceskip=\fontdimen2\font plus
\BIBentryALTinterwordstretchfactor\fontdimen3\font minus
  \fontdimen4\font\relax}
\providecommand{\BIBforeignlanguage}[2]{{%
\expandafter\ifx\csname l@#1\endcsname\relax
\typeout{** WARNING: IEEEtranS.bst: No hyphenation pattern has been}%
\typeout{** loaded for the language `#1'. Using the pattern for}%
\typeout{** the default language instead.}%
\else
\language=\csname l@#1\endcsname
\fi
#2}}
\providecommand{\BIBdecl}{\relax}
\BIBdecl

\bibitem{agha2014firm}
A.-A. Agha-Mohammadi, S.~Chakravorty, and N.~M. Amato, ``{FIRM}: Sampling-based
  feedback motion-planning under motion uncertainty and imperfect
  measurements,'' \emph{International Journal of Robotics Research}, vol.~33,
  no.~2, pp. 268--304, 2014.

\bibitem{agha2014health}
A.-a. Agha-mohammadi, N.~K. Ure, J.~P. How, and J.~Vian, ``Health aware
  stochastic planning for persistent package delivery missions using
  quadrotors,'' in \emph{IEEE/RSJ International Conference on Intelligent
  Robots and Systems (IROS)}, 2014, pp. 3389--3396.

\bibitem{ahmadi2019safe}
M.~{Ahmadi}, A.~{Singletary}, J.~W. {Burdick}, and A.~D. {Ames}, ``Safe policy
  synthesis in multi-agent {POMDP}s via discrete-time barrier functions,'' in
  \emph{IEEE Conference on Decision and Control (CDC)}, 2019, pp. 4797--4803.

\bibitem{amato2016policy}
C.~Amato, G.~Konidaris, A.~Anders, G.~Cruz, J.~P. How, and L.~P. Kaelbling,
  ``Policy search for multi-robot coordination under uncertainty,''
  \emph{International Journal of Robotics Research}, vol.~35, no.~14, pp.
  1760--1778, 2016.

\bibitem{armstrong2007oracular}
N.~Armstrong-Crews and M.~Veloso, ``Oracular partially observable {Markov}
  decision processes: A very special case,'' in \emph{IEEE International
  Conference on Robotics and Automation (ICRA)}, 2007, pp. 2477--2482.

\bibitem{astrom1965optimal}
K.~J. {\AA}strom, ``Optimal control of {Markov} processes with incomplete state
  information,'' \emph{Journal of Mathematical Analysis and Applications},
  vol.~10, no.~1, pp. 174--205, 1965.

\bibitem{atanasov2014nonmyopic}
N.~Atanasov, B.~Sankaran, J.~Le~Ny, G.~J. Pappas, and K.~Daniilidis,
  ``Nonmyopic view planning for active object classification and pose
  estimation,'' \emph{IEEE Transactions on Robotics}, vol.~30, no.~5, pp.
  1078--1090, 2014.

\bibitem{atrash2009bayesian}
A.~Atrash and J.~Pineau, ``A {B}ayesian reinforcement learning approach for
  customizing human-robot interfaces,'' in \emph{International Conference on
  Intelligent User Interfaces ({IUI})}, 2009, pp. 355--360.

\bibitem{atrash2010bayesian}
------, ``A bayesian method for learning {POMDP} observation parameters for
  robot interaction management systems,'' in \emph{The {POMDP} practitioners
  workshop}, 2010.

\bibitem{bai2015intention}
H.~Bai, S.~Cai, N.~Ye, D.~Hsu, and W.~S. Lee, ``Intention-aware online {POMDP}
  planning for autonomous driving in a crowd,'' in \emph{IEEE International
  Conference on Robotics and Automation (ICRA)}, 2015, pp. 454--460.

\bibitem{bai2012unmanned}
H.~Bai, D.~Hsu, M.~J. Kochenderfer, and W.~S. Lee, ``Unmanned aircraft
  collision avoidance using continuous-state {POMDPs},'' in \emph{Robotics:
  Science and Systems}, 2011.

\bibitem{bai2014integrated}
H.~Bai, D.~Hsu, and W.~S. Lee, ``Integrated perception and planning in the
  continuous space: A {POMDP} approach,'' \emph{International Journal of
  Robotics Research}, vol.~33, no.~9, pp. 1288--1302, 2014.

\bibitem{bai2010monte}
H.~Bai, D.~Hsu, W.~S. Lee, and V.~A. Ngo, ``Monte {C}arlo value iteration for
  continuous-state {POMDP}s,'' in \emph{9th International Workshop on the
  Algorithmic Foundations of Robotics ({WAFR})}, 2010, pp. 175--191.

\bibitem{BaiHsu13a}
H.~Bai, D.~Hsu, and W.~Lee, ``Planning how to learn,'' in \emph{IEEE
  International Conference on Robotics and Automation (ICRA)}, 2013.

\bibitem{balaban2019unifying}
E.~Balaban, S.~B. Johnson, and M.~J. Kochenderfer, ``Unifying system health
  management and automated decision making,'' \emph{Journal of Artificial
  Intelligence Research}, vol.~65, pp. 487--518, 2019.

\bibitem{bellman1966dynamic}
R.~Bellman, ``Dynamic programming,'' \emph{Science}, vol. 153, no. 3731, pp.
  34--37, 1966.

\bibitem{bernstein2002complexity}
D.~S. Bernstein, R.~Givan, N.~Immerman, and S.~Zilberstein, ``The complexity of
  decentralized control of {Markov} decision processes,'' \emph{Mathematics of
  operations research}, vol.~27, no.~4, pp. 819--840, 2002.

\bibitem{bertsekas2005dynamic}
D.~P. Bertsekas, ``Dynamic programming and suboptimal control: {A} survey from
  {ADP} to {MPC},'' \emph{European Journal of Control}, vol.~11, no. 4-5, pp.
  310--334, 2005.

\bibitem{best2019dec}
G.~Best, O.~M. Cliff, T.~Patten, R.~R. Mettu, and R.~Fitch, ``{Dec-MCTS}:
  Decentralized planning for multi-robot active perception,''
  \emph{International Journal of Robotics Research}, vol.~38, no. 2-3, pp.
  316--337, 2019.

\bibitem{bhattacharya2020multiagent}
S.~Bhattacharya, S.~Kailas, S.~Badyal, S.~Gil, and D.~Bertsekas, ``Multiagent
  rollout and policy iteration for {POMDP} with application to multi-robot
  repair problems,'' in \emph{Conference on Robot Learning}, 2020.

\bibitem{bohg2014grasping}
J.~Bohg, A.~Morales, T.~Asfour, and D.~Kragic, ``Data-driven grasp
  synthesis—a survey,'' \emph{IEEE Transactions on Robotics}, vol.~30, no.~2,
  pp. 289--309, 2014.

\bibitem{boonvisut2012estimation}
P.~Boonvisut and M.~C. {\c{C}}avu{\c{s}}o{\u{g}}lu, ``Estimation of soft tissue
  mechanical parameters from robotic manipulation data,'' \emph{IEEE/ASME
  Transactions on Mechatronics}, vol.~18, no.~5, pp. 1602--1611, 2012.

\bibitem{brechtel2013solving}
S.~Brechtel, T.~Gindele, and R.~Dillmann, ``Solving continuous {POMDP}s: Value
  iteration with incremental learning of an efficient space representation,''
  in \emph{International Conference on Machine Learning (ICML)}, 2013, pp.
  370--378.

\bibitem{brechtel2014probabilistic}
------, ``Probabilistic decision-making under uncertainty for autonomous
  driving using continuous {POMDP}s,'' in \emph{IEEE International Conference
  on Intelligent Transportation Systems (ITSC)}, 2014, pp. 392--399.

\bibitem{browne2012survey}
C.~B. Browne, E.~Powley, D.~Whitehouse, S.~M. Lucas, P.~I. Cowling,
  P.~Rohlfshagen, S.~Tavener, D.~Perez, S.~Samothrakis, and S.~Colton, ``A
  survey of {M}onte {C}arlo tree search methods,'' \emph{IEEE Transactions on
  Computational Intelligence and AI in Games}, vol.~4, no.~1, pp. 1--43, 2012.

\bibitem{broz2011designing}
F.~Broz, I.~Nourbakhsh, and R.~Simmons, ``Designing {POMDP} models of socially
  situated tasks,'' in \emph{IEEE International Symposium on Robot and Human
  Interactive Communication (RO-MAN)}, 2011, pp. 39--46.

\bibitem{brunskill2008continuous}
E.~Brunskill, L.~P. Kaelbling, T.~Lozano-Perez, and N.~Roy, ``Continuous-state
  {POMDP}s with hybrid dynamics,'' in \emph{International Symposium on
  Artificial Intelligence and Mathematics (ISAIM)}, 2008.

\bibitem{buffet2009factored}
O.~Buffet and D.~Aberdeen, ``The factored policy-gradient planner,''
  \emph{Artificial Intelligence}, vol. 173, no. 5-6, pp. 722--747, 2009.

\bibitem{burks2019optimal}
L.~Burks, I.~Loefgren, and N.~R. Ahmed, ``Optimal continuous state {POMDP}
  planning with semantic observations: A variational approach,'' \emph{IEEE
  Transactions on Robotics}, vol.~35, no.~6, pp. 1488--1507, 2019.

\bibitem{cai2020hyp}
P.~Cai, Y.~Luo, D.~Hsu, and W.~S. Lee, ``{HyP-DESPOT}: A hybrid parallel
  algorithm for online planning under uncertainty,'' \emph{International
  Journal of Robotics Research}, vol.~40, no. 2-3, 2021.

\bibitem{cao2021tare}
C.~Cao, H.~Zhu, H.~Choset, and J.~Zhang, ``{TARE}: A hierarchical framework for
  efficiently exploring complex {3D} environments,'' in \emph{Robotics: Science
  and Systems}, 2021.

\bibitem{capitan2013decentralized}
J.~Capitan, M.~T. Spaan, L.~Merino, and A.~Ollero, ``Decentralized multi-robot
  cooperation with auctioned {POMDP}s,'' \emph{International Journal of
  Robotics Research}, vol.~32, no.~6, pp. 650--671, 2013.

\bibitem{cassandra1997incremental}
A.~Cassandra, M.~L. Littman, and N.~L. Zhang, ``Incremental pruning: A simple,
  fast, exact method for partially observable {Markov} decision processes,'' in
  \emph{Proc. 13th Conference in Uncertainty in Artificial Intelligence (UAI)},
  1997, pp. 54--61.

\bibitem{cassandra1994acting}
A.~R. Cassandra, L.~P. Kaelbling, and M.~L. Littman, ``Acting optimally in
  partially observable stochastic domains,'' in \emph{Proc. 12th National
  Conference on Artificial Intelligence (AAAI)}, 1994, pp. 1023--1028.

\bibitem{charnes1963deterministic}
A.~Charnes and W.~W. Cooper, ``Deterministic equivalents for optimizing and
  satisficing under chance constraints,'' \emph{Operations Research}, vol.~11,
  no.~1, pp. 18--39, 1963.

\bibitem{chaudhari2013sampling}
P.~Chaudhari, S.~Karaman, D.~Hsu, and E.~Frazzoli, ``Sampling-based algorithms
  for continuous-time {POMDP}s,'' in \emph{American Control Conference (ACC)},
  2013, pp. 4604--4610.

\bibitem{chen2016pomdp}
M.~Chen, E.~Frazzoli, D.~Hsu, and W.~S. Lee, ``{POMDP}-lite for robust robot
  planning under uncertainty,'' in \emph{IEEE International Conference on
  Robotics and Automation (ICRA)}, 2016, pp. 5427--5433.

\bibitem{chen2018planning}
M.~Chen, S.~Nikolaidis, H.~Soh, D.~Hsu, and S.~Srinivasa, ``Planning with trust
  for human-robot collaboration,'' in \emph{ACM/IEEE International Conference
  on Human-Robot Interaction (HRI)}, 2018, pp. 307--315.

\bibitem{chen2003bayesian}
Z.~Chen, ``{Bayesian filtering: From Kalman filters to particle filters, and
  beyond},'' \emph{Statistics}, vol. 182, no.~1, pp. 1--69, 2003.

\bibitem{corah2019distributed}
M.~Corah and N.~Michael, ``Distributed matroid-constrained submodular
  maximization for multi-robot exploration: theory and practice,''
  \emph{Autonomous Robots}, vol.~43, no.~2, pp. 485--501, 2019.

\bibitem{cunningham2015mpdm}
A.~G. Cunningham, E.~Galceran, R.~M. Eustice, and E.~Olson, ``{MPDM}:
  Multipolicy decision-making in dynamic, uncertain environments for autonomous
  driving,'' in \emph{IEEE International Conference on Robotics and Automation
  (ICRA)}, 2015, pp. 1670--1677.

\bibitem{dallaire2009bayesian}
P.~Dallaire, C.~Besse, S.~Ross, and B.~Chaib-draa, ``Bayesian reinforcement
  learning in continuous {POMDP}s with {G}aussian processes,'' in
  \emph{IEEE/RSJ International Conference on Intelligent Robots and Systems
  (IROS)}, 2009, pp. 2604--2609.

\bibitem{thiebaux2016rao}
P.~H. de~Rodrigues Quemel~e Assis~Santana, S.~Thi{\'{e}}baux, and B.~C.
  Williams, ``{RAO*}: An algorithm for chance-constrained {POMDP}'s,'' in
  \emph{AAAI Conference on Artificial Intelligence}, 2016, pp. 3308--3314.

\bibitem{dempster1977maximum}
A.~P. Dempster, N.~M. Laird, and D.~B. Rubin, ``Maximum likelihood from
  incomplete data via the {EM} algorithm,'' \emph{Journal of the Royal
  Statistical Society: Series B (Methodological)}, vol.~39, no.~1, pp. 1--22,
  1977.

\bibitem{durrantwhyte2006slam}
H.~F. Durrant{-}Whyte and T.~Bailey, ``Simultaneous localization and mapping:
  part {I},'' \emph{{IEEE} Robotics and Automation Magazine}, vol.~13, no.~2,
  pp. 99--110, 2006.

\bibitem{eidenberger2010active}
R.~Eidenberger and J.~Scharinger, ``Active perception and scene modeling by
  planning with probabilistic {6D} object poses,'' in \emph{IEEE/RSJ
  International Conference on Intelligent Robots and Systems (IROS)}, 2010, pp.
  1036--1043.

\bibitem{filliat2003map}
D.~Filliat and J.-A. Meyer, ``Map-based navigation in mobile robots:: I. a
  review of localization strategies,'' \emph{Cognitive Systems Research},
  vol.~4, no.~4, pp. 243--282, 2003.

\bibitem{foka2003predictive}
A.~F. Foka and P.~E. Trahanias, ``Predictive control of robot velocity to avoid
  obstacles in dynamic environments,'' in \emph{IEEE/RSJ International
  Conference on Intelligent Robots and Systems (IROS)}, 2003, pp. 370--375.

\bibitem{garg2019learning}
N.~P. Garg, D.~Hsu, and W.~S. Lee, ``Learning to grasp under uncertainty using
  {POMDPs},'' in \emph{IEEE International Conference on Robotics and Automation
  (ICRA)}.\hskip 1em plus 0.5em minus 0.4em\relax IEEE, 2019, pp. 2751--2757.

\bibitem{garrett2020}
C.~R. Garrett, C.~Paxton, T.~Lozano-Pérez, L.~P. Kaelbling, and D.~Fox,
  ``Online replanning in belief space for partially observable task and motion
  problems,'' in \emph{IEEE International Conference on Robotics and Automation
  (ICRA)}, 2020, pp. 5678--5684.

\bibitem{ghavamzadeh2015bayesian}
M.~Ghavamzadeh, S.~Mannor, J.~Pineau, and A.~Tamar, ``Bayesian reinforcement
  learning: A survey,'' \emph{Foundations and Trends in Machine Learning},
  vol.~8, no. 5-6, pp. 359--483, 2015.

\bibitem{goldhoorn2018searching}
A.~Goldhoorn, A.~Garrell, R.~Alqu{\'e}zar, and A.~Sanfeliu, ``Searching and
  tracking people with cooperative mobile robots,'' \emph{Autonomous Robots},
  vol.~42, no.~4, pp. 739--759, 2018.

\bibitem{Grady2013AutomatedMA}
D.~K. Grady, M.~Moll, and L.~E. Kavraki, ``Automated model approximation for
  robotic navigation with {POMDP}s,'' \emph{IEEE International Conference on
  Robotics and Automation (ICRA)}, pp. 78--84, 2013.

\bibitem{grady2015extending}
------, ``Extending the applicability of {POMDP} solutions to robotic tasks,''
  \emph{IEEE Transactions on Robotics}, vol.~31, no.~4, pp. 948--961, 2015.

\bibitem{sorin2019deep}
S.~Grigorescu, B.~Trasnea, T.~Cocias, and G.~Macesanu, ``A survey of deep
  learning techniques for autonomous driving,'' \emph{Journal of Field
  Robotics}, vol.~37, no.~3, pp. 362--386, 2020.

\bibitem{gupta2017decision}
A.~Gupta, D.~Bessonov, and P.~Li, ``A decision-theoretic approach to
  detection-based target search with a {UAV},'' in \emph{IEEE/RSJ International
  Conference on Intelligent Robots and Systems (IROS)}, 2017, pp. 5304--5309.

\bibitem{hafner2019learning}
D.~Hafner, T.~P. Lillicrap, I.~Fischer, R.~Villegas, D.~Ha, H.~Lee, and
  J.~Davidson, ``Learning latent dynamics for planning from pixels,'' in
  \emph{International Conference on Machine Learning, {ICML}}, 2019, pp.
  2555--2565.

\bibitem{hauskrecht2000value}
M.~Hauskrecht, ``Value-function approximations for partially observable
  {Markov} decision processes,'' \emph{Journal of Artificial Intelligence
  Research}, vol.~13, pp. 33--94, 2000.

\bibitem{hoelscher2018utilizing}
J.~Hoelscher, D.~Koert, J.~Peters, and J.~Pajarinen, ``Utilizing human feedback
  in {POMDP} execution and specification,'' in \emph{IEEE-RAS 18th
  International Conference on Humanoid Robots}, 2018, pp. 104--111.

\bibitem{hoey2010automated}
J.~Hoey, P.~Poupart, A.~von Bertoldi, T.~Craig, C.~Boutilier, and
  A.~Mihailidis, ``Automated handwashing assistance for persons with dementia
  using video and a partially observable {Markov} decision process,''
  \emph{Computer Vision and Image Understanding}, vol. 114, no.~5, pp.
  503--519, 2010.

\bibitem{hollinger2009efficient}
G.~Hollinger, S.~Singh, J.~Djugash, and A.~Kehagias, ``Efficient multi-robot
  search for a moving target,'' \emph{International Journal of Robotics
  Research}, vol.~28, no.~2, pp. 201--219, 2009.

\bibitem{horowitz2013interactive}
M.~Horowitz and J.~Burdick, ``Interactive non-prehensile manipulation for
  grasping via {POMDP}s,'' in \emph{IEEE International Conference on Robotics
  and Automation (ICRA)}, 2013, pp. 3257--3264.

\bibitem{hsiao2007grasping}
K.~Hsiao, L.~P. Kaelbling, and T.~Lozano-Perez, ``Grasping {POMDP}s,'' in
  \emph{{IEEE} International Conference on Robotics and Automation (ICRA)},
  2007, pp. 4685--4692.

\bibitem{hsiao2008robust}
K.~Hsiao, T.~Lozano-P{\'e}rez, and L.~P. Kaelbling, ``Robust belief-based
  execution of manipulation programs,'' in \emph{8th International Workshop on
  the Algorithmic Foundations of Robotics ({WAFR})}, 2008.

\bibitem{ychsu2020}
Y.~Hsu, S.~Gopalswamy, S.~Saripalli, and D.~A. Shell, ``A {POMDP} treatment of
  vehicle-pedestrian interaction: Implicit coordination via uncertainty-aware
  planning,'' in \emph{IEEE/RSJ International Conference on Intelligent Robots
  and Systems (IROS)}, 2020, pp. 1984--1991.

\bibitem{hubmann2017decision}
C.~Hubmann, M.~Becker, D.~Althoff, D.~Lenz, and C.~Stiller, ``Decision making
  for autonomous driving considering interaction and uncertain prediction of
  surrounding vehicles,'' in \emph{IEEE Intelligent Vehicles Symposium (IV)},
  2017, pp. 1671--1678.

\bibitem{indelman2015planning}
V.~Indelman, L.~Carlone, and F.~Dellaert, ``Planning in the continuous domain:
  {A} generalized belief space approach for autonomous navigation in unknown
  environments,'' \emph{International Journal of Robotics Research}, vol.~34,
  no.~7, pp. 849--882, 2015.

\bibitem{jaulmes2007formal}
R.~Jaulmes, J.~Pineau, and D.~Precup, ``A formal framework for robot learning
  and control under model uncertainty,'' in \emph{IEEE International Conference
  on Robotics and Automation (ICRA)}, 2007, pp. 2104--2110.

\bibitem{kaelbling1998planning}
L.~P. Kaelbling, M.~L. Littman, and A.~R. Cassandra, ``Planning and acting in
  partially observable stochastic domains,'' \emph{Artificial Intelligence},
  vol. 101, no. 1-2, pp. 99--134, 1998.

\bibitem{kaelbling2017pre}
L.~P. Kaelbling and T.~Lozano-P{\'e}rez, ``Pre-image backchaining in belief
  space for mobile manipulation,'' in \emph{15th International Symposium on
  Robotics Research (ISRR)}.\hskip 1em plus 0.5em minus 0.4em\relax Springer,
  2011, pp. 383--400.

\bibitem{kaelbling2013integrated}
------, ``Integrated task and motion planning in belief space,''
  \emph{International Journal of Robotics Research}, vol.~32, no. 9-10, pp.
  1194--1227, 2013.

\bibitem{KarMa19}
P.~Karkus, X.~Ma, D.~Hsu, L.~Kaelbling, W.~Lee, and T.~Lozano-Perez,
  ``Differentiable algorithm networks for composable robot learning,'' in
  \emph{Proc. Robotics: Science \& Systems}, 2019.

\bibitem{kavraki1996probabilistic}
L.~E. Kavraki, P.~Svestka, J.-C. Latombe, and M.~H. Overmars, ``Probabilistic
  roadmaps for path planning in high-dimensional configuration spaces,''
  \emph{IEEE Transactions on Robotics and Automation}, vol.~12, no.~4, pp.
  566--580, 1996.

\bibitem{kim2021review}
D.~Kim, S.-H. Kim, T.~Kim, B.~B. Kang, M.~Lee, W.~Park, S.~Ku, D.~Kim, J.~Kwon,
  H.~Lee, J.~Bae, Y.-L. Park, K.-J. Cho, and S.~Jo, ``Review of machine
  learning methods in soft robotics,'' \emph{PLOS ONE}, vol.~16, no.~2, pp.
  1--24, 2021.

\bibitem{kim2021plgrim}
S.-K. Kim, A.~Bouman, G.~Salhotra, D.~D. Fan, K.~Otsu, J.~Burdick, and A.-a.
  Agha-mohammadi, ``{PLGRIM}: Hierarchical value learning for large-scale
  exploration in unknown environments,'' in \emph{Proceedings of the
  International Conference on Automated Planning and Scheduling (ICAPS)},
  vol.~31, 2021, pp. 652--662.

\bibitem{klimenko2014tapir}
D.~Klimenko, J.~Song, and H.~Kurniawati, ``{TAPIR}: A software toolkit for
  approximating and adapting {POMDP} solutions online,'' in \emph{Proc.
  Australasian Conference on Robotics and Automation}, 2014.

\bibitem{kober2013reinforcement}
J.~Kober, J.~A. Bagnell, and J.~Peters, ``Reinforcement learning in robotics: A
  survey,'' \emph{International Journal of Robotics Research}, vol.~32, no.~11,
  pp. 1238--1274, 2013.

\bibitem{koenig1998xavier}
S.~Koenig and R.~Simmons, \emph{Xavier: A robot navigation architecture based
  on partially observable markov decision process models}.\hskip 1em plus 0.5em
  minus 0.4em\relax MIT Press, 1998, pp. 91--122.

\bibitem{kopitkov2017no}
D.~Kopitkov and V.~Indelman, ``No belief propagation required: Belief space
  planning in high-dimensional state spaces via factor graphs, the matrix
  determinant lemma, and re-use of calculation,'' \emph{International Journal
  of Robotics Research}, vol.~36, no.~10, pp. 1088--1130, 2017.

\bibitem{korsah2013comprehensive}
G.~A. Korsah, A.~Stentz, and M.~B. Dias, ``A comprehensive taxonomy for
  multi-robot task allocation,'' \emph{International Journal of Robotics
  Research}, vol.~32, no.~12, pp. 1495--1512, 2013.

\bibitem{koval2016pre}
M.~C. Koval, N.~S. Pollard, and S.~S. Srinivasa, ``Pre-and post-contact policy
  decomposition for planar contact manipulation under uncertainty,''
  \emph{International Journal of Robotics Research}, vol.~35, no. 1-3, pp.
  244--264, 2016.

\bibitem{kurniawati2008sarsop}
H.~Kurniawati, D.~Hsu, and W.~S. Lee, ``{SARSOP}: Efficient point-based {POMDP}
  planning by approximating optimally reachable belief spaces,'' in
  \emph{Robotics: Science and Systems}, 2008.

\bibitem{kurniawati2013online}
H.~Kurniawati and V.~Yadav, ``An online {POMDP} solver for uncertainty planning
  in dynamic environment,'' in \emph{16th International Symposium on Robotics
  Research (ISRR)}, 2013, pp. 611--629.

\bibitem{lagoudakis2003least}
M.~G. Lagoudakis and R.~Parr, ``Least-squares policy iteration,'' \emph{Journal
  of Machine Learning Research}, vol.~4, pp. 1107--1149, 2003.

\bibitem{lasota2017safe}
P.~A. Lasota, T.~Fong, and J.~A. Shah, ``A survey of methods for safe
  human-robot interaction,'' \emph{Foundations and Trends in Robotics}, vol.~5,
  no.~4, pp. 261--349, 2017.

\bibitem{lauri2017multi}
M.~Lauri, E.~Hein{\"a}nen, and S.~Frintrop, ``Multi-robot active information
  gathering with periodic communication,'' in \emph{IEEE International
  Conference on Robotics and Automation (ICRA)}, 2017, pp. 851--856.

\bibitem{lauri2020multib}
M.~Lauri and F.~A. Oliehoek, ``Multi-agent active perception with prediction
  rewards,'' in \emph{Advances in Neural Information Processing Systems 33},
  2020.

\bibitem{lauri2020multi}
M.~Lauri, J.~Pajarinen, and J.~Peters, ``Multi-agent active information
  gathering in discrete and continuous-state decentralized {POMDP}s by policy
  graph improvement,'' \emph{Autonomous Agents and Multi-Agent Systems},
  vol.~34, no.~2, 2020.

\bibitem{lauri2016planning}
M.~Lauri and R.~Ritala, ``Planning for robotic exploration based on forward
  simulation,'' \emph{Robotics and Autonomous Systems}, vol.~83, pp. 15--31,
  2016.

\bibitem{lee2018monte}
J.~Lee, G.-H. Kim, P.~Poupart, and K.-E. Kim, ``{Monte-Carlo Tree Search for
  Constrained POMDPs},'' in \emph{Advances in Neural Information Processing
  Systems 31}, 2018, pp. 7934--7943.

\bibitem{lee2021}
Y.~Lee, P.~Cai, and D.~Hsu, ``{MAGIC:} learning macro-actions for online
  {POMDP} planning,'' in \emph{Robotics: Science and Systems}, 2021.

\bibitem{li2016act}
J.~K. Li, D.~Hsu, and W.~S. Lee, ``{Act to see and see to act: POMDP planning
  for objects search in clutter},'' in \emph{IEEE/RSJ International Conference
  on Intelligent Robots and Systems (IROS)}, 2016, pp. 5701--5707.

\bibitem{littman1995learning}
M.~L. Littman, A.~R. Cassandra, and L.~P. Kaelbling, ``Learning policies for
  partially observable environments: {Scaling} up,'' in \emph{International
  Conference on Machine Learning (ICML)}, 1995, pp. 362--370.

\bibitem{long2018towards}
P.~Long, T.~Fanl, X.~Liao, W.~Liu, H.~Zhang, and J.~Pan, ``Towards optimally
  decentralized multi-robot collision avoidance via deep reinforcement
  learning,'' in \emph{IEEE International Conference on Robotics and Automation
  (ICRA)}, 2018, pp. 6252--6259.

\bibitem{lovejoy1991survey}
W.~S. Lovejoy, ``A survey of algorithmic methods for partially observed
  {Markov} decision processes,'' \emph{Annals of Operations Research}, vol.~28,
  no.~1, pp. 47--65, 1991.

\bibitem{luo2018autonomous}
Y.~Luo, P.~Cai, A.~Bera, D.~Hsu, W.~S. Lee, and D.~Manocha, ``{PORCA}: Modeling
  and planning for autonomous driving among many pedestrians,'' \emph{IEEE
  Robotics and Automation Letters}, vol.~3, no.~4, pp. 3418--3425, 2018.

\bibitem{martinez2009bayesian}
R.~Martinez-Cantin, N.~de~Freitas, E.~Brochu, J.~Castellanos, and A.~Doucet,
  ``A {Bayesian} exploration-exploitation approach for optimal online sensing
  and planning with a visually guided mobile robot,'' \emph{Autonomous Robots},
  vol.~27, no.~2, pp. 93--103, 2009.

\bibitem{matignon2012coordinated}
L.~Matignon, L.~Jeanpierre, and A.-I. Mouaddib, ``Coordinated multi-robot
  exploration under communication constraints using decentralized {Markov}
  decision processes,'' in \emph{{AAAI} Conference on Artificial Intelligence},
  2012.

\bibitem{miller2009pomdp}
S.~A. Miller, Z.~A. Harris, and E.~K. Chong, ``A {POMDP} framework for
  coordinated guidance of autonomous {UAV}s for multitarget tracking,''
  \emph{EURASIP Journal on Advances in Signal Processing}, vol. 2009, no.
  724597, 2009.

\bibitem{monahan1982state}
G.~E. Monahan, ``State of the art—a survey of partially observable {Markov}
  decision processes: theory, models, and algorithms,'' \emph{Management
  science}, vol.~28, no.~1, pp. 1--16, 1982.

\bibitem{monso2012pomdp}
P.~Mons{\'o}, G.~Aleny{\`a}, and C.~Torras, ``{POMDP} approach to robotized
  clothes separation,'' in \emph{IEEE/RSJ International Conference on
  Intelligent Robots and Systems (IROS)}, 2012, pp. 1324--1329.

\bibitem{moon1996expectation}
T.~K. Moon, ``The expectation-maximization algorithm,'' \emph{IEEE Signal
  processing magazine}, vol.~13, no.~6, pp. 47--60, 1996.

\bibitem{Munos14}
R.~Munos, ``From bandits to {M}onte-{C}arlo tree search: The optimistic
  principle applied to optimization and planning,'' \emph{Foundations and
  Trends in Machine Learning}, vol.~7, no.~1, pp. 1--129, 2014.

\bibitem{Nguyen20}
H.~Nguyen, B.~Daley, X.~Song, C.~Amato, and R.~Platt, ``Belief-grounded
  networks for accelerated robot learning under partial observability,'' in
  \emph{Conference on Robot Learning}, 2020, pp. 1640--1653.

\bibitem{oliehoek2016concise}
F.~A. Oliehoek and C.~Amato, \emph{A concise introduction to decentralized
  POMDPs}.\hskip 1em plus 0.5em minus 0.4em\relax Springer, 2016.

\bibitem{omidshafiei2015decentralized}
S.~Omidshafiei, A.~Agha{-}mohammadi, C.~Amato, and J.~P. How, ``Decentralized
  control of partially observable {Markov} decision processes using belief
  space macro-actions,'' in \emph{IEEE International Conference on Robotics and
  Automation (ICRA)}, 2015, pp. 5962--5969.

\bibitem{omidshafiei2017decentralized}
S.~Omidshafiei, A.-A. Agha-Mohammadi, C.~Amato, S.-Y. Liu, J.~P. How, and
  J.~Vian, ``Decentralized control of multi-robot partially observable {M}arkov
  decision processes using belief space macro-actions,'' \emph{International
  Journal of Robotics Research}, vol.~36, no.~2, pp. 231--258, 2017.

\bibitem{ong2010planning}
S.~C. Ong, S.~W. Png, D.~Hsu, and W.~S. Lee, ``Planning under uncertainty for
  robotic tasks with mixed observability,'' \emph{International Journal of
  Robotics Research}, vol.~29, no.~8, pp. 1053--1068, 2010.

\bibitem{pajarinen2014robotic}
J.~Pajarinen and V.~Kyrki, ``Robotic manipulation in object composition
  space,'' in \emph{IEEE/RSJ International Conference on Intelligent Robots and
  Systems (IROS)}, 2014, pp. 1--6.

\bibitem{pajarinen2017robotic}
------, ``Robotic manipulation of multiple objects as a {POMDP},''
  \emph{Artificial Intelligence}, vol. 247, pp. 213--228, 2017.

\bibitem{pajarinen2020pomdp}
J.~Pajarinen, J.~Lundell, and V.~Kyrki, ``{POMDP} planning under object
  composition uncertainty: Application to robotic manipulation,'' \emph{IEEE
  Transactions on Robotics}, pp. 1--16, 2022.

\bibitem{peltzer2022fig}
O.~Peltzer, A.~Bouman, S.~Kim, R.~Senanayake, J.~Ott, H.~Delecki, M.~Sobue,
  M.~J. Kochenderfer, M.~Schwager, J.~Burdick, and A.~Agha{-}mohammadi,
  ``{FIG-OP:} exploring large-scale unknown environments on a fixed time
  budget,'' \emph{CoRR}, vol. abs/2203.06316, 2022.

\bibitem{petric2019hierarchical}
F.~Petric and Z.~Kova{\v{c}}i{\'c}, ``Hierarchical {POMDP} framework for a
  robot-assisted {ASD} diagnostic protocol,'' in \emph{ACM/IEEE International
  Conference on Human-Robot Interaction (HRI)}, 2019, pp. 286--293.

\bibitem{phiquepal2019}
C.~Phiquepal and M.~Toussaint, ``Combined task and motion planning under
  partial observability: An optimization-based approach,'' in \emph{IEEE
  International Conference on Robotics and Automation (ICRA)}, 2019, pp.
  9000--9006.

\bibitem{pineau2003point}
J.~Pineau, G.~Gordon, S.~Thrun \emph{et~al.}, ``Point-based value iteration: An
  anytime algorithm for {POMDP}s,'' in \emph{Proc. 18th International Joint
  Conference on Artificial Intelligence (IJCAI)}, vol.~3, 2003, pp. 1025--1032.

\bibitem{pineau2006anytime}
J.~Pineau, G.~Gordon, and S.~Thrun, ``Anytime point-based approximations for
  large {POMDP}s,'' \emph{Journal of Artificial Intelligence Research},
  vol.~27, pp. 335--380, 2006.

\bibitem{pineau2003towards}
J.~Pineau, M.~Montemerlo, M.~Pollack, N.~Roy, and S.~Thrun, ``Towards robotic
  assistants in nursing homes: Challenges and results,'' \emph{Robotics and
  Autonomous Systems}, vol.~42, no. 3-4, pp. 271--281, 2003.

\bibitem{pomerleau1988alvinn}
D.~A. Pomerleau, ``{ALVINN: An autonomous land vehicle in a neural network},''
  in \emph{Advances in Neural Information Processing Systems 1}, 1988, pp.
  305--313.

\bibitem{poupart2005exploiting}
P.~Poupart, ``Exploiting structure to efficiently solve large scale partially
  observable {Markov} decision processes,'' Ph.D. dissertation, University of
  Toronto, 2005.

\bibitem{pusse2019hybrid}
F.~Pusse and M.~Klusch, ``Hybrid online {POMDP} planning and deep reinforcement
  learning for safer self-driving cars,'' in \emph{IEEE Intelligent Vehicles
  Symposium (IV)}, 2019, pp. 1013--1020.

\bibitem{puterman1994markov}
M.~L. Puterman, \emph{Markov Decision Processes: Discrete Stochastic Dynamic
  Programming}, 1st~ed.\hskip 1em plus 0.5em minus 0.4em\relax John Wiley \&
  Sons, Inc., 1994.

\bibitem{ragi2014decentralized}
S.~Ragi and E.~K. Chong, ``Decentralized guidance control of {UAV}s with
  explicit optimization of communication,'' \emph{Journal of Intelligent \&
  Robotic Systems}, vol.~73, no.~1, pp. 811--822, 2014.

\bibitem{ramachandran2019personalized}
A.~Ramachandran, S.~S. Sebo, and B.~Scassellati, ``{Personalized robot tutoring
  using the assistive tutor POMDP (AT-POMDP)},'' in \emph{AAAI Conference on
  Artificial Intelligence}, 2019, pp. 8050--8057.

\bibitem{rasmussen2006gaussian}
C.~Rasmussen and C.~Williams, \emph{Gaussian Processes for Machine
  Learning}.\hskip 1em plus 0.5em minus 0.4em\relax MIT Press, 2006.

\bibitem{rosen2020mixed}
E.~Rosen, D.~Whitney, M.~Fishman, D.~Ullman, and S.~Tellex, ``Mixed reality as
  a bidirectional communication interface for human-robot interaction,'' in
  \emph{IEEE/RSJ International Conference on Intelligent Robots and Systems
  (IROS)}, 2020, pp. 11\,431--11\,438.

\bibitem{rosenthal2011modeling}
S.~Rosenthal and M.~Veloso, ``Modeling humans as observation providers using
  {POMDPs},'' in \emph{IEEE International Symposium on Robot and Human
  Interactive Communication (RO-MAN)}, 2011, pp. 53--58.

\bibitem{ross2008bayesian}
S.~Ross, B.~Chaib-draa, and J.~Pineau, ``Bayesian reinforcement learning in
  continuous {POMDPs} with application to robot navigation,'' in \emph{IEEE
  International Conference on Robotics and Automation (ICRA)}, 2008, pp.
  2845--2851.

\bibitem{ross2008online}
S.~Ross, J.~Pineau, S.~Paquet, and B.~Chaib-Draa, ``Online planning algorithms
  for {POMDP}s,'' \emph{Journal of Artificial Intelligence Research}, vol.~32,
  pp. 663--704, 2008.

\bibitem{satsangi2018exploiting}
Y.~Satsangi, S.~Whiteson, F.~A. Oliehoek, and M.~T. Spaan, ``Exploiting
  submodular value functions for scaling up active perception,''
  \emph{Autonomous Robots}, vol.~42, no.~2, pp. 209--233, 2018.

\bibitem{schmidt2010learning}
S.~R. Schmidt-Rohr, M.~L{\"o}sch, and R.~Dillmann, ``Learning flexible,
  multi-modal human-robot interaction by observing human-human-interaction,''
  in \emph{19th International Symposium in Robot and Human Interactive
  Communication}, 2010, pp. 582--587.

\bibitem{schrittwieser2020mastering}
J.~Schrittwieser, I.~Antonoglou, T.~Hubert, K.~Simonyan, L.~Sifre, S.~Schmitt,
  A.~Guez, E.~Lockhart, D.~Hassabis, T.~Graepel \emph{et~al.}, ``{Mastering
  Atari, Go, chess and shogi by planning with a learned model},''
  \emph{Nature}, vol. 588, no. 7839, pp. 604--609, 2020.

\bibitem{shani2013survey}
G.~Shani, J.~Pineau, and R.~Kaplow, ``A survey of point-based {POMDP}
  solvers,'' \emph{Autonomous Agents and Multi-Agent Systems}, vol.~27, no.~1,
  pp. 1--51, 2013.

\bibitem{silver2010monte}
D.~Silver and J.~Veness, ``Monte-carlo planning in large {POMDP}s,'' in
  \emph{Advances in Neural Information Processing Systems 23}, 2010, pp.
  2164--2172.

\bibitem{slade2017simultaneous}
P.~Slade, P.~Culbertson, Z.~Sunberg, and M.~Kochenderfer, ``Simultaneous active
  parameter estimation and control using sampling-based {B}ayesian
  reinforcement learning,'' in \emph{IEEE/RSJ International Conference on
  Intelligent Robots and Systems (IROS)}, 2017, pp. 804--810.

\bibitem{smallwood1973optimal}
R.~D. Smallwood and E.~J. Sondik, ``The optimal control of partially observable
  {Markov} processes over a finite horizon,'' \emph{Operations Research},
  vol.~21, no.~5, pp. 1071--1088, 1973.

\bibitem{smith2004heuristic}
T.~Smith and R.~Simmons, ``Heuristic search value iteration for {POMDP}s,'' in
  \emph{Proc. 20th Conference in Uncertainty in Artificial Intelligence (UAI)},
  2004, pp. 520--527.

\bibitem{smith2005point}
------, ``Point-based {POMDP} algorithms: Improved analysis and
  implementation,'' in \emph{Proc. 21st Conference in Uncertainty in Artificial
  Intelligence (UAI)}, 2005, p. 542–549.

\bibitem{somani2013despot}
A.~Somani, N.~Ye, D.~Hsu, and W.~S. Lee, ``{DESPOT}: Online {POMDP} planning
  with regularization,'' in \emph{Advances in Neural Information Processing
  Systems 26}, 2013, pp. 1772--1780.

\bibitem{sondik1971optimal}
E.~Sondik, ``The optimal control of partially observable {Markov} processes,''
  Ph.D. dissertation, Stanford University, 1971.

\bibitem{song2016intention}
W.~Song, G.~Xiong, and H.~Chen, ``Intention-aware autonomous driving
  decision-making in an uncontrolled intersection,'' \emph{Mathematical
  Problems in Engineering}, vol. 2016, 2016.

\bibitem{spaan2008cooperative}
M.~T. Spaan, ``Cooperative active perception using {POMDP}s,'' in \emph{AAAI
  Workshop on Advancements in {POMDP} Solvers}, 2008.

\bibitem{spaan2010active}
M.~T. Spaan, T.~S. Veiga, and P.~U. Lima, ``Active cooperative perception in
  network robot systems using {POMDP}s,'' in \emph{IEEE/RSJ International
  Conference on Intelligent Robots and Systems (IROS)}, 2010, pp. 4800--4805.

\bibitem{spaan2005perseus}
M.~T. Spaan and N.~Vlassis, ``Perseus: Randomized point-based value iteration
  for {POMDP}s,'' \emph{Journal of Artificial Intelligence Research}, vol.~24,
  pp. 195--220, 2005.

\bibitem{sridharan2010planning}
M.~Sridharan, J.~Wyatt, and R.~Dearden, ``Planning to see: A hierarchical
  approach to planning visual actions on a robot using {POMDP}s,''
  \emph{Artificial Intelligence}, vol. 174, no.~11, pp. 704--725, 2010.

\bibitem{sukkar2019multi}
F.~Sukkar, G.~Best, C.~Yoo, and R.~Fitch, ``Multi-robot region-of-interest
  reconstruction with {Dec-MCTS},'' in \emph{IEEE International Conference on
  Robotics and Automation (ICRA)}, 2019, pp. 9101--9107.

\bibitem{sunberg2017value}
Z.~N. Sunberg, C.~J. Ho, and M.~J. Kochenderfer, ``The value of inferring the
  internal state of traffic participants for autonomous freeway driving,'' in
  \emph{American Control Conference (ACC)}, 2017, pp. 3004--3010.

\bibitem{sunberg2018online}
Z.~N. Sunberg and M.~J. Kochenderfer, ``Online algorithms for {{POMDP}}s with
  continuous state, action, and observation spaces,'' in \emph{Proc. 28th
  International Conference on Automated Planning and Scheduling {ICAPS}}, 2018,
  pp. 259--263.

\bibitem{sutton1999between}
R.~S. Sutton, D.~Precup, and S.~Singh, ``{Between MDPs and semi-MDPs: A
  framework for temporal abstraction in reinforcement learning},''
  \emph{Artificial intelligence}, vol. 112, no. 1-2, pp. 181--211, 1999.

\bibitem{szer2005maastar}
D.~Szer, F.~Charpillet, and S.~Zilberstein, ``{MAA*}: A heuristic search
  algorithm for solving decentralized {POMDP}s,'' in \emph{Proc. 21st
  Conference in Uncertainty in Artificial Intelligence (UAI)}, 2005, p.
  576–583.

\bibitem{taha2011pomdp}
T.~Taha, J.~V. Mir{\'o}, and G.~Dissanayake, ``A {POMDP} framework for
  modelling human interaction with assistive robots,'' in \emph{IEEE
  International Conference on Robotics and Automation (ICRA)}, 2011, pp.
  544--549.

\bibitem{thornton2018autonomous}
S.~Thornton, ``Autonomous vehicle speed control for safe navigation of occluded
  pedestrian crosswalk,'' \emph{arXiv preprint arXiv:1802.06314}, 2018.

\bibitem{ulbrich2015towards}
S.~Ulbrich and M.~Maurer, ``Towards tactical lane change behavior planning for
  automated vehicles,'' in \emph{IEEE 18th International Conference on
  Intelligent Transportation Systems (ITSC)}, 2015, pp. 989--995.

\bibitem{undurti2010online}
A.~Undurti and J.~P. How, ``{An online algorithm for constrained POMDPs},'' in
  \emph{IEEE International Conference on Robotics and Automation (ICRA)}, 2010,
  pp. 3966--3973.

\bibitem{unhelkar2019semi}
V.~V. Unhelkar, S.~Li, and J.~A. Shah, ``Semi-supervised learning of
  decision-making models for human-robot collaboration,'' in \emph{Conference
  on Robot Learning (CoRL)}, 2019, pp. 192--203.

\bibitem{unhelkar2020decision}
------, ``Decision-making for bidirectional communication in sequential
  human-robot collaborative tasks,'' in \emph{ACM/IEEE International Conference
  on Human-Robot Interaction (HRI)}, 2020, pp. 329--341.

\bibitem{valimaki2016optimizing}
T.~V{\"a}lim{\"a}ki and R.~Ritala, ``Optimizing gaze direction in a visual
  navigation task,'' in \emph{IEEE International Conference on Robotics and
  Automation (ICRA)}, 2016, pp. 1427--1432.

\bibitem{van2012efficient}
J.~Van Den~Berg, S.~Patil, and R.~Alterovitz, ``Efficient approximate value
  iteration for continuous {G}aussian {POMDP}s,'' in \emph{AAAI Conference on
  Artificial Intelligence}, 2012.

\bibitem{vanegas2016uav}
F.~Vanegas, D.~Campbell, M.~Eich, and F.~Gonzalez, ``{UAV} based target finding
  and tracking in {GPS}-denied and cluttered environments,'' in \emph{IEEE/RSJ
  International Conference on Intelligent Robots and Systems (IROS)}, 2016, pp.
  2307--2313.

\bibitem{viseras2019deepig}
A.~Viseras and R.~Garcia, ``{DeepIG}: Multi-robot information gathering with
  deep reinforcement learning,'' \emph{IEEE Robotics and Automation Letters},
  vol.~4, no.~3, pp. 3059--3066, 2019.

\bibitem{wandzel2019multi}
A.~Wandzel, Y.~Oh, M.~Fishman, N.~Kumar, W.~L. LS, and S.~Tellex,
  ``Multi-object search using object-oriented {POMDPs},'' in \emph{IEEE
  International Conference on Robotics and Automation (ICRA)}, 2019, pp.
  7194--7200.

\bibitem{Wang20model}
R.~E. Wang, J.~C. Kew, D.~Lee, T.~E. Lee, T.~Zhang, B.~Ichter, J.~Tan, and
  A.~Faust, ``Model-based reinforcement learning for decentralized multiagent
  rendezvous,'' in \emph{Conference on Robot Learning}, 2020, pp. 711--725.

\bibitem{wang2020pomp}
Y.~Wang, F.~Giuliari, R.~Berra, A.~Castellini, A.~Del~Bue, A.~Farinelli,
  M.~Cristani, and F.~Setti, ``{POMP: POMCP-based Online Motion Planning for
  active visual search in indoor environments},'' in \emph{British Machine
  Vision Conference}, 2020.

\bibitem{wang2018bounded}
Y.~Wang, S.~Chaudhuri, and L.~E. Kavraki, ``{Bounded Policy Synthesis for
  POMDPs with Safe-Reachability Objectives},'' in \emph{Proceedings of the 17th
  International Conference on Autonomous Agents and Multi-Agent Systems
  (AAMAS)}, 2018, pp. 238--246.

\bibitem{wang2017anticipatory}
Z.~Wang, A.~Boularias, K.~M{\"u}lling, B.~Sch{\"o}lkopf, and J.~Peters,
  ``Anticipatory action selection for human--robot table tennis,''
  \emph{Artificial Intelligence}, vol. 247, pp. 399--414, 2017.

\bibitem{wei2011point}
J.~Wei, J.~M. Dolan, J.~M. Snider, and B.~Litkouhi, ``A point-based {MDP} for
  robust single-lane autonomous driving behavior under uncertainties,'' in
  \emph{IEEE International Conference on Robotics and Automation (ICRA)}, 2011,
  pp. 2586--2592.

\bibitem{white1991survey}
C.~C. White, ``A survey of solution techniques for the partially observed
  {Markov} decision process,'' \emph{Annals of Operations Research}, vol.~32,
  no.~1, pp. 215--230, 1991.

\bibitem{wray2021}
K.~H. Wray, B.~Lange, A.~Jamgochian, S.~J. Witwicki, A.~Kobashi,
  S.~Hagaribommanahalli, and D.~Ilstrup, ``{POMDPs} for safe visibility
  reasoning in autonomous vehicles,'' in \emph{IEEE International Conference on
  Intelligence and Safety for Robotics (ISR)}, 2021, pp. 191--195.

\bibitem{wray2017}
K.~H. Wray, S.~J. Witwicki, and S.~Zilberstein, ``Online decision-making for
  scalable autonomous systems,'' in \emph{International Joint Conference on
  Artificial Intelligence ({IJCAI})}, 2017, pp. 4768--4774.

\bibitem{Xiao20}
Y.~Xiao, J.~Hoffman, T.~Xia, and C.~Amato, ``Learning multi-robot decentralized
  macro-action-based policies via a centralized {Q}-net,'' in \emph{IEEE
  International Conference on Robotics and Automation (ICRA)}, 2020, pp.
  10\,695--10\,701.

\bibitem{xiao2019online}
Y.~Xiao, S.~Katt, A.~ten Pas, S.~Chen, and C.~Amato, ``Online planning for
  target object search in clutter under partial observability,'' in \emph{IEEE
  International Conference on Robotics and Automation (ICRA)}, 2019, pp.
  8241--8247.

\bibitem{xu2016review}
J.~Xu and H.-S. Yoon, ``A review on mechanical and hydraulic system modeling of
  excavator manipulator system,'' \emph{Journal of Construction Engineering},
  vol. 2016, 2016.

\bibitem{yi2019indoor}
S.~Yi, C.~Nam, and K.~Sycara, ``Indoor pursuit-evasion with hybrid hierarchical
  partially observable {M}arkov decision processes for multi-robot systems,''
  in \emph{14th International Symposium on Distributed Autonomous Robotic
  Systems}, 2018, pp. 251--264.

\bibitem{zhang2017role}
H.~Zhang, J.~Chen, H.~Fang, and L.~Dou, ``A role-based {POMDP}s approach for
  decentralized implicit cooperation of multiple agents,'' in \emph{IEEE
  International Conference on Control \& Automation}, 2017, pp. 496--501.

\bibitem{zhang2013active}
S.~Zhang, M.~Sridharan, and C.~Washington, ``Active visual planning for mobile
  robot teams using hierarchical {POMDP}s,'' \emph{IEEE Transactions on
  Robotics}, vol.~29, no.~4, pp. 975--985, 2013.

\bibitem{zheng2021}
K.~Zheng, Y.~Sung, G.~Konidaris, and S.~Tellex, ``Multi-resolution {POMDP}
  planning for multi-object search in {3D},'' in \emph{IEEE/RSJ International
  Conference on Intelligent Robots and Systems (IROS)}, 2021, pp. 2022--2029.

\bibitem{zheng2018pomdp}
W.~Zheng, B.~Wu, and H.~Lin, ``{POMDP} model learning for human robot
  collaboration,'' in \emph{IEEE Conference on Decision and Control
  (CDC)}.\hskip 1em plus 0.5em minus 0.4em\relax IEEE, 2018, pp. 1156--1161.

\bibitem{zhou2017probabilistic}
J.~Zhou, R.~Paolini, A.~M. Johnson, J.~A. Bagnell, and M.~T. Mason, ``A
  probabilistic planning framework for planar grasping under uncertainty,''
  \emph{{IEEE} Robotics and Automation Letters}, vol.~2, no.~4, pp. 2111--2118,
  2017.

\end{thebibliography}

\begin{IEEEbiography}[{\includegraphics[width=1in,height=1.25in,clip,keepaspectratio]{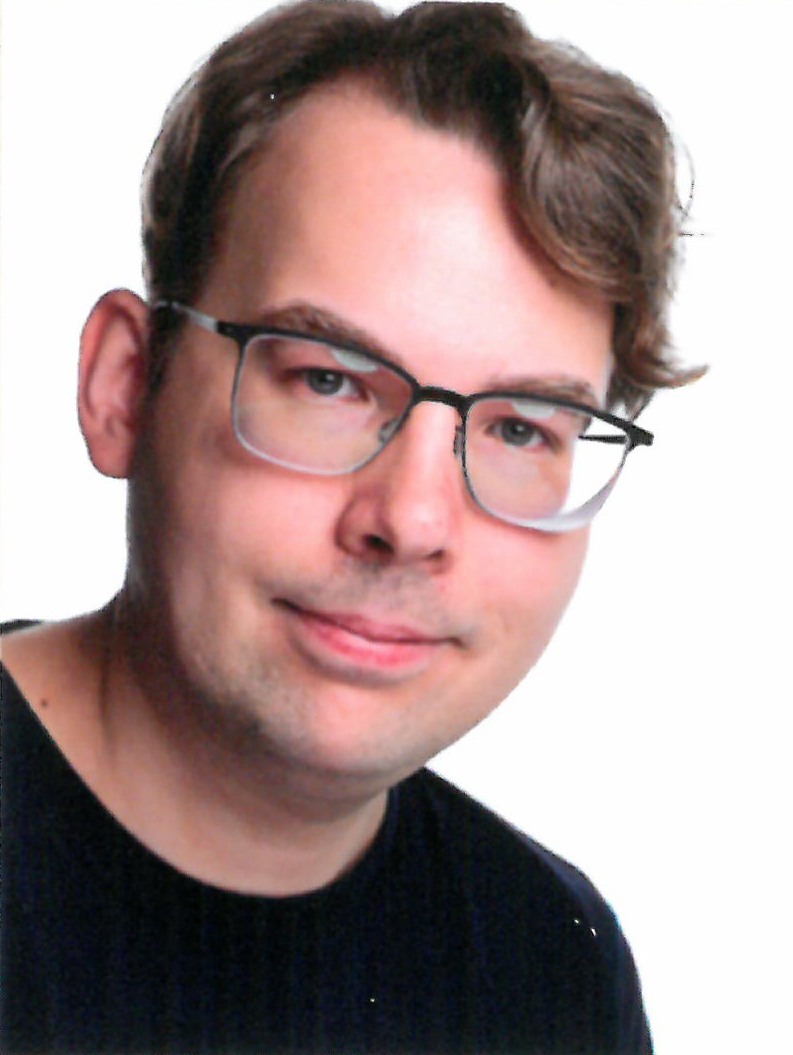}}]{Mikko Lauri}
is a post-doctoral researcher at the Department of Informatics at University of Hamburg, Germany.
He received the M.Sc. degree in electrical engineering and the doctoral degree in automation science from Tampere University of Technology, Finland, in 2010 and 2016, respectively.
Mikko's research interests include single-agent and multi-agent decision-making under uncertainty, active perception, and computer vision for robotics.
\end{IEEEbiography}

\begin{IEEEbiography}[{\includegraphics[width=1in,height=1.25in,clip,keepaspectratio]{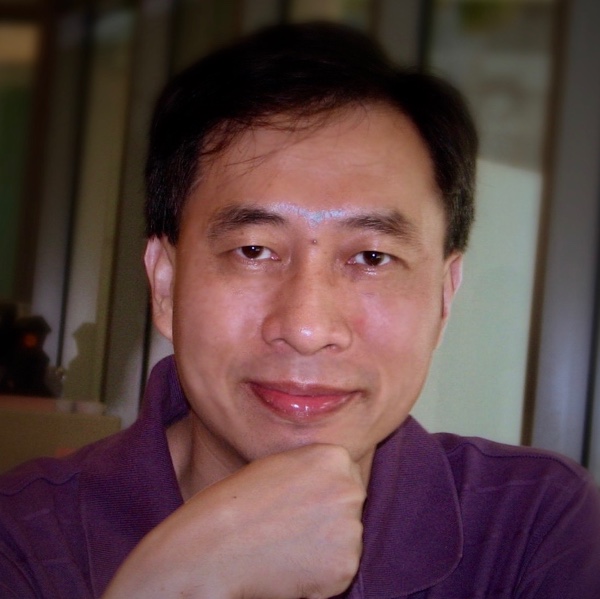}}]{David Hsu}
is a Provost's Chair Professor at the National University of Singapore (NUS) and the director of Smart Systems Institute. He received the B.Sc. degree in computer science \& mathematics from the University of British Columbia and a doctoral degree in computer science from Stanford University. He co-founded the NUS Advanced Robotics Center in 2013. He founded the NUS AI Laboratory in 2019 and served as the founding director. He held visiting positions at MIT Aeronautics \& Astronautics Department and at CMU Robotics Institute.

David's research interests span robot planning and learning under uncertainty, human-robot collaboration, and computational structural biology. His work has won several international awards, including the Test of Time Award at Robotics: Science \& Systems, 2021.
He has chaired or co-chaired several major international robotics conferences, and served on the editorial boards of IEEE Transactions on Robotics and Journal of Artificial Intelligence Research. He is an associate editor of the International Journal of Robotics Research. 
\end{IEEEbiography}

\begin{IEEEbiography}[{\includegraphics[width=1in,height=1.25in,clip,keepaspectratio]{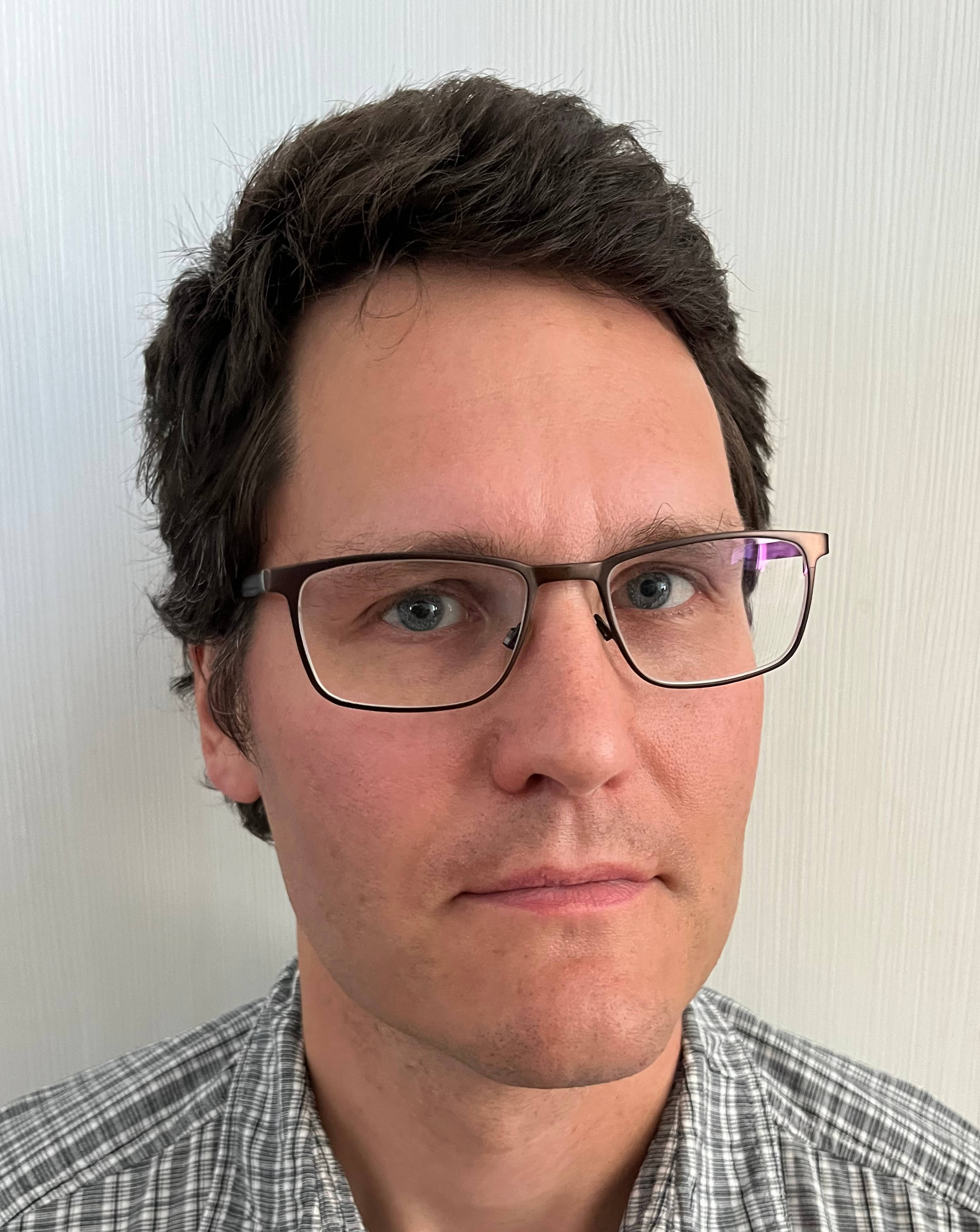}}]{Joni Pajarinen}
is an Assistant Professor at Aalto University where he is leading from 2020 the Aalto Robot Learning research group that develops new decision making and control methods and applies them to novel robotic tasks. Joni received the doctoral degree in computer science from Aalto University, Finland in 2013. From 2013 to 2016 he was a post-doctoral researcher focusing on robotics at Aalto University. From 2016 to 2022 he was a post-doctoral researcher / research group leader at the Intelligent Autonomous Systems institute at TU Darmstadt, Germany.
His research interests include reinforcement learning, robotics, planning under uncertainty, multi-agent decision making, and mobile manipulation.
\end{IEEEbiography}

\end{document}